\documentclass[12pt, letterpaper]{article}

\usepackage{times}

\usepackage{graphicx}
\usepackage{tabularx}
\usepackage{array}
\usepackage{longtable}
\usepackage{booktabs}
\usepackage{ragged2e}
\usepackage{tikz}
\usetikzlibrary{positioning, arrows.meta, shapes.geometric, calc, backgrounds, fit}
\usepackage{url}
\usepackage{pdflscape}

\newcolumntype{Y}{>{\RaggedRight\arraybackslash}X}
\newcolumntype{P}[1]{>{\RaggedRight\arraybackslash}p{#1}}
\usepackage[round,authoryear]{natbib}
\setcitestyle{aysep={,}}

\usepackage{caption}
\renewcommand{\baselinestretch}{2.0}\normalsize

\makeatletter
\renewcommand\section{\@startsection {section}{1}{\z@}%
                                   {-3.5ex \@plus -1ex \@minus -.2ex}%
                                   {2.3ex \@plus.2ex}%
                                   {\normalfont\normalsize\bfseries}}
\renewcommand\subsection{\@startsection {subsection}{2}{\z@}%
                                      {-3.25ex \@plus -1ex \@minus -.2ex}%
                                      {1.5ex \@plus .2ex}%
                                      {\normalfont\normalsize\bfseries}}
\makeatother

\title{\textbf{Redrawing the AI Map: A Theory of Accountability Boundaries in Agentic Ecosystems}}
\date{}

\begin{document}

{\renewcommand{\baselinestretch}{1.0}\normalsize
\begin{center}
{\LARGE\bfseries Redrawing the AI Map: A Theory of Accountability Boundaries in Agentic Ecosystems\par}
\vspace{1.2em}
{\normalsize
Muhammad Zia Hydari\\
University of Pittsburgh\\
\texttt{hydari@alum.mit.edu}\\
ORCID: 0000-0003-4522-326X\par}
\vspace{0.6em}
{\normalsize
Farooq Muzaffar\\
Ensi.ai\\
\texttt{fmuzaffar@alum.mit.edu}\\
ORCID: 0009-0007-2310-8144\par}
\end{center}
\vspace{1em}
}

\begin{center}\textbf{Abstract}\end{center}
{\renewcommand{\baselinestretch}{1.15}\normalsize
\noindent Agentic AI orchestrators reduce the interface and assembly costs of composing information systems capabilities across organizational boundaries, apparently accelerating the modular forms that digital innovation theory leads us to expect. Yet AI-enabled capabilities whose outputs require evidence, review, signoff, or assignable responsibility may retain integrated accountability boundaries even when their technical interfaces become modular. We introduce \textit{accountability assets}: complementary assets whose strategic value derives from making AI-supported outputs legitimate, auditable, reviewable, and assignable to a responsible party. When verification is costly and responsibility is difficult to transfer, accountability assets limit the boundary-shifting effects of agentic orchestration, even when technical decomposability is high. We further introduce \textit{rule debt}: the latent governance burden that accrues when organizational decision rules migrate from formal information systems into ungoverned agentic execution environments, producing unmanaged policy artifacts that must eventually be inventoried, versioned, tested, and reconciled. The theory integrates digital innovation, transaction cost, complementary-assets, digital platform governance, and IS control perspectives to develop seven propositions explaining how agentic reductions in interface and assembly costs, verification cost, responsibility transferability, and accountability-asset cospecialization shape three capability-level boundary strategies: component, integrated, and dual-track. The propositions further explain how appropriability conditions accountability-asset value capture, how orchestrator intent capture enables value appropriation, and how boundary misconfiguration produces rule debt. Structured theoretical illustrations across document processing, legal services, audit, clinical decision support, and procurement discipline the boundary logic. The theory advances IS scholarship by specifying when digital modularization extends to organizational disaggregation and when it does not, by extending IS control theory to agentic settings in which decision rules execute outside governed information systems, and by identifying accountability-asset cospecialization as a structural determinant of digital platform governance asymmetries.\par}

\vspace{1em}

{\renewcommand{\baselinestretch}{1.0}\normalsize
\par}

\newpage
\renewcommand{\baselinestretch}{2.0}\normalsize
\section{INTRODUCTION}

Agentic AI orchestrators are reshaping how information systems capabilities are assembled across organizational boundaries. An agentic orchestrator interprets a user request, identifies relevant complements, routes work among tools, monitors intermediate outputs, and assembles a workflow that previously required either managerial direction inside a firm or substantial integration effort by the customer. Digital innovation research gives strong reasons to expect that modular technical architectures should loosen coupling among components, expand ecosystems, and redistribute influence toward architectural control points \citep{yoo2010,tiwana2010,henfridsson2013}. Agentic orchestration intensifies this expectation by making the assembly of components cheaper for users than at any prior point in the history of enterprise information systems. A coordination-cost reading of the Coasian mechanism therefore predicts pressure toward vertical disintegration as market assembly becomes cheaper \citep{coase1937,williamson1985}. This prediction is useful but incomplete: it explains why execution can move across technical interfaces, not whether the accountability boundary can move with it. AI-enabled capabilities whose outputs require evidence, review, signoff, or assignable responsibility may retain integrated accountability boundaries even when their technical interfaces become modular.

This paper explains the puzzle by shifting attention from the interface boundary to the accountability boundary. The interface boundary is the location of user interaction, task initiation, and workflow routing. The accountability boundary is the location of responsibility, evidence, review, signoff, and post hoc defensibility. Agentic orchestration can move the first boundary without moving the second. When AI-enabled outputs must be defensible, reviewable, legitimate, and assignable to a responsible party, the strategically relevant assets are \textit{accountability assets}: complementary assets whose value derives from making outputs legitimate, auditable, reviewable, and assignable after the fact.

The motivating phenomenon is the strategic response emerging among vertical AI firms, in which a firm cedes workflow and interface to a general-purpose agent and exposes specialized domain functionality as a callable capability. Practitioners refer to this as ``going headless,'' retaining the back-end specialized capabilities while ceding the user-facing interface layer to the orchestrator. The pattern is empirically vivid but is not the theoretical unit of analysis. The unit of analysis in this paper is the AI-enabled capability and its accountability boundary across vertical firms, agentic orchestrators, and customer organizations. The same firm may componentize document extraction, retain integration around audit signoff, and operate a dual-track boundary strategy around legal drafting and filed legal commitments. The question is not whether a firm goes headless in general, but whether a specific AI-enabled capability is best governed through a component, integrated, or dual-track boundary strategy.

The theory turns on two primary dimensions. \textit{Verification cost} captures the cost of determining whether an AI-supported output is correct, complete, defensible, and usable in organizational action. \textit{Responsibility transferability} captures the degree to which responsibility for an output can be contractually, professionally, or operationally shifted across organizational boundaries. When verification cost is low and responsibility is transferable, agentic orchestration is likely to produce a component strategy. When verification cost is high or responsibility is difficult to transfer, the relevant organization retains control over accountability assets through an integrated strategy. When a capability bundle contains separable edge functions that are componentizable but an accountability-bearing core that is not, it adopts a dual-track strategy: edge functions are exposed as components while the core remains integrated. The integrated core may still be reachable through bounded modular interfaces for initiation, input submission, status retrieval, or delivery, but control over evidence, review, lineage, signoff, and responsibility remains inside the accountability boundary. The boundary is therefore determined by the accountability regime of the capability or sub-capability, not by interface modularity alone.

The theory makes three contributions to information systems scholarship and is intended as a Theory Development article: it develops new constructs, elaborates boundary conditions for digital modularization, and integrates multiple theoretical perspectives into an IS theory of agentic boundary strategy. First, it qualifies digital innovation theory's expectations about modularization. Prior work suggests that modular architectures and recombinant logics produce more decentralized organizing forms as interface and assembly costs fall \citep{yoo2010,henfridsson2013,lyytinen2016}. The theory developed here specifies a structural condition under which this expectation fails: technical decomposability does not entail organizational decomposability when accountability assets are cospecialized with the focal capability and when responsibility cannot transfer across the technical interface.

Second, the theory extends IS control and IT governance theory to agentic settings. IS control research has theorized portfolios of formal and informal controls that align IS-related work with organizational objectives \citep{kirsch1996,kirsch1997,choudhury2003,wiener2016}. IT governance research has specified the allocation of decision rights over IT-enabled activities \citep{sambamurthy1999,weill2004,tiwana2014}. When an agent executes decision rules outside the governed information systems where rules have historically lived, the resulting policy fragments accumulate as \textit{rule debt}: a customer-side governance burden including inventorying, ownership, versioning, testing, drift monitoring, and reconciliation with formal policy. Rule debt names a customer-side governance burden that arises when decision rules execute outside the information systems that previously absorbed them.

Third, the theory extends digital platform governance theory by identifying cospecialization with accountability assets as a structural determinant of complementor vulnerability that has not previously been specified \citep{tiwana2010,parker2017,constantinides2018}. Agentic orchestrators direct routing, ranking, discovery, default selection, evaluation, and substitution among complements even when the formal interface is open. Complementors whose capabilities are cospecialized with accountability assets are less vulnerable to envelopment because the orchestrator must reproduce not only the task output but also the defensibility infrastructure that makes the output usable.

The propositions generalize beyond agentic AI to settings in which modularization pressure meets accountability constraints, including platform-mediated professional services, regulated digital services, and knowledge-intensive work organized around defensible decisions. Agentic AI is theoretically revealing because it makes the divergence between the interface boundary and accountability boundary unusually visible.

The remainder of the paper proceeds as follows. Section 2 reviews digital innovation theory, transaction cost economics, complementary assets and appropriability, digital platform governance, and IS control as the theoretical foundations on which the contribution rests. In \citet{gregor2006} terms, the resulting theory is primarily a theory for explaining and predicting boundary strategies, although it also yields design implications for agentic information systems. Section 3 explains how agentic orchestration reduces interface and assembly costs and creates pressure toward disaggregation while leaving orchestrator power intact. Section 4 defines accountability assets and shows why they constrain disaggregation. Section 5 develops a theory of boundary choice. Section 6 states seven propositions with derivations, mechanisms, alternative predictions, and scope conditions. Section 7 presents the theoretical model. Section 8 provides structured theoretical illustrations. Section 9 discusses implications for IS theory, including digital innovation, IS control, IT governance, and digital platform research. Section 10 addresses limitations and future research.

\section{THEORETICAL BACKGROUND}

\subsection{Digital Modularization and the Boundary Effects of Digital Innovation}

A central claim of digital innovation theory is that digital technologies enable modular, layered architectures that loosen the historical coupling between technical components and organizational actors. This expectation is consistent with modularity research showing how decomposable architectures permit variation and recombination across components \citep{baldwin2000}, and with digital infrastructure research emphasizing layered architectures, generativity, and distributed innovation across organizational boundaries \citep{tilson2010,yoo2010}. \citet{yoo2010} argued that digital innovation reorganizes work around recombinable digital objects whose distributed, generative properties allow innovation to occur across organizational boundaries. \citet{henfridsson2013} traced the generative mechanisms through which digital infrastructures evolve, emphasizing that architectural openness invites participation by actors whose contributions were not anticipated by the original designers. \citet{lyytinen2016} showed that digital innovation networks exhibit organizing logics distinct from traditional firm-bounded innovation. These contributions generate a baseline expectation: as digital technologies reduce the cost of recombining capabilities across organizational boundaries, organizing forms should shift toward more disaggregated, ecosystem-based configurations.

Agentic AI orchestration intensifies this pressure. Where prior generations of digital architecture required customers to identify, integrate, and orchestrate components themselves, agentic orchestrators perform much of the assembly work autonomously: search costs fall because the orchestrator identifies available complements, workflow-assembly costs fall because it composes workflows, monitoring costs fall because it observes intermediate outputs, and adaptation costs fall because it revises plans when steps fail. The digital innovation tradition has recognized that technical modularity and organizational disaggregation are not equivalent \citep{henfridsson2013,constantinides2018}, but it has not fully specified the conditions under which technical decomposability fails to translate into organizational decomposition in agentic AI settings. The accountability-assets framework developed in this paper addresses that gap by identifying a class of socio-technical assets that remain bound to organizational responsibility even when the technical interface becomes callable.

\subsection{Firm Boundaries and Transaction Costs}

Transaction cost economics provides the second foundation. \citet{coase1937} argued that firms arise because using markets is costly; \citet{williamson1985} developed this insight through bounded rationality, opportunism, uncertainty, and asset specificity, showing that integration becomes attractive when market exchange exposes parties to hazards that contracts cannot resolve. Property-rights theory reinforced this logic by showing that residual control over assets shapes ex ante investment incentives when contracts are incomplete \citep{grossman1986,hart1990}, and holdup risk emerges when relationship-specific investments create appropriable quasi-rents \citep{klein1978}.

Agentic orchestration elaborates the classical comparison in two ways. First, it lowers the cost of using markets, which the framework treats as the right reason to shift outward. Second, it makes more visible a distinction that is easy to collapse in ordinary sourcing decisions: an AI-supported output is not only a unit of work that can be sourced internally or externally, but also a candidate organizational action that may need to be evaluated, defended, and stood behind. A transaction-cost comparison focused on market assembly and bilateral contractual hazards therefore leaves under-specified where defensibility resides when execution, orchestration, and accountable use are split across actors. Agentic orchestration creates triadic settings in which a platform actor stands between buyer and seller and the buyer may retain accountability for the output regardless of which seller produced it. The technically critical asset (the model, the tool) and the accountability-critical asset (the evidence and review infrastructure) may sit in different organizations, so the boundary of task execution and the boundary of accountability can diverge. This divergence is a substantive boundary condition that requires theoretical attention.

\subsection{Complementary Assets and Appropriability}

\citet{teece1986} showed that profits from innovation often accrue not to the inventor alone, but to the party that controls the complementary assets required to commercialize the innovation. Complementary assets include distribution, manufacturing, service, brand, regulatory approval, customer relationships, and other resources that transform an invention into a product or service. Teece distinguished generic assets from specialized and cospecialized assets, with cospecialized assets exhibiting two-way dependence between the innovation and the asset. The relevance of complementary assets increases when appropriability is weak: if an innovation can be imitated, value capture depends less on owning the innovation itself and more on controlling the assets required to make it valuable in use. The framework has been productively extended to digital and platform settings \citep{jacobides2018}, but the focus has remained on commercialization.

Many AI-enabled capabilities face weak appropriability because general-purpose models can approximate classification, extraction, summarization, drafting, and first-pass reasoning. Where appropriability is weak, control over cospecialized complementary assets becomes the principal route to durable value capture. The framework was developed primarily for tangible commercialization assets that convert inventions into usable products, but AI-supported work introduces a different kind of cospecialized asset. Accountability assets do not commercialize an innovation; they make its outputs usable in organizational settings in which someone must defend the decision. A model output that is technically excellent but cannot be defended is not commercially equivalent to one that is technically adequate and defensible. Teece's framework predicts the importance of complementary assets in general but does not specify the strategic significance of assets whose role is post hoc defensibility rather than ex ante commercialization. This paper extends the framework to identify accountability assets as a theoretically distinct subset of complementary assets that often become cospecialized with the AI-enabled capability they support.

\subsection{Digital Platform Governance and Envelopment}

Platform scholarship distinguishes the architecture of participation from the control of value capture. \citet{tiwana2010} developed the foundational framework of platform architecture and governance, arguing that platform evolution is shaped by the coevolution of architectural decisions, governance arrangements, and environmental dynamics. Subsequent IS research has elaborated this framework: \citet{constantinides2018} emphasized that platforms and infrastructures evolve through interactions between technical architecture and organizational governance, \citet{tiwana2014} developed a comprehensive treatment of platform governance including decision rights, control mechanisms, and pricing structures, and \citet{parker2017} showed that platform ecosystems invert the firm by enabling value creation outside organizational boundaries while preserving platform governance authority. Related work on platform envelopment \citep{eisenmann2011} and platform-complementor competition shows that platform owners can absorb, re-rank, or substitute complementor functionality even when technical access remains open \citep{zhu2018}.

These literatures converge on a recurring theme: technical interface openness does not entail economic openness. Open platform strategy distinguishes granting access from devolving control \citep{boudreau2010}. Agentic orchestrators occupy the platform position in a new and particularly consequential way. They control the user-facing surface, the routing logic among complements, the ranking of tool choices, the substitution of first-party for third-party functionality, and the learning loop that determines which capabilities attract repeated demand. A complementor that connects through an open protocol still depends on the orchestrator for discovery, invocation, and user access. The classical asset-specificity logic \citep{klein1978} reappears in a new form: relationship-specific investments in a single orchestrator's routing, ranking, evaluation, and user experience create appropriable quasi-rents that the orchestrator can capture once the investments are sunk.

Existing platform governance theory has been less developed for the question of how accountability for outputs interacts with platform power. A complementor that depends on a platform for discovery is exposed; a complementor that depends on a platform for discovery and also surrenders the assets that make its outputs defensible faces a compounded exposure. This paper extends platform governance theory by identifying accountability-asset cospecialization as a structural determinant of complementor vulnerability, alongside the architectural and governance variables that the literature has previously emphasized.

\subsection{IS Control, IT Governance, and Algorithmic Accountability}

A fourth literature stream is essential for the theory developed here: IS control and IT governance. Kirsch (1996, 1997) established the foundational framework for analyzing how organizations exercise authority over information systems development and use, identifying formal (behavior, outcome) and informal (clan, self) modes of control. \citet{choudhury2003} extended this framework to outsourced settings, showing that control portfolios evolve over time. \citet{wiener2016} integrated and expanded the IS control framework, emphasizing the configuration and enactment of controls across project stakeholders. IT governance research has specified how decision rights over IT-enabled activities should be allocated across organizational actors \citep{sambamurthy1999,weill2004}. These literatures share a common substrate: rules, controls, decisions, and authority flow through identifiable organizational arrangements, including information systems, that make oversight and accountability possible. The information system itself absorbs much of the governance work: approval workflows enforce policy, audit logs preserve evidence, permission structures encode authority, workflow systems route exceptions to accountable owners, and master data and policy repositories anchor the organization's official position.

Agentic AI complicates this substrate. When an agent interprets a customer organization's intent and executes a workflow through external complements, the rules that previously lived inside governed information systems can migrate into informal agent instructions. A prompt or a natural-language directive can specify when to escalate a claim, classify a transaction, summarize a medical message, rank a supplier, or draft a legal response. At small scale, this is convenient. At organizational scale, it creates rules that exist outside the formal control configuration that IS control theory has analyzed, specifically rules that are ungoverned in a specific sense: they have no formal owner, no version history, no test record, no drift monitoring, and no reconciliation against authoritative policy. The customer organization retains responsibility for the decisions these rules produce but has lost the IS-mediated controls that previously made those decisions governable. Recent IS work on algorithmic organizing and AI management has begun to address related concerns \citep{faraj2018,berente2021}, and the framework developed in this paper extends these conversations by identifying a specific governance phenomenon, rule debt, that emerges when agentic AI reconfigures the boundary between governed information systems and the organizational decisions those systems previously absorbed.

Accountability itself has been theorized as a relationship in which an actor must explain and justify conduct to a forum that can question, judge, and impose consequences \citep{bovens2007}. In information systems work, accountability is enacted through the design, configuration, and use of information systems. Professional and regulated settings illustrate the point. Audit standards require documentation sufficient to support review and conclusions \citep{pcaob_as1215}, while audit evidence standards emphasize the reliability, source, and nature of evidence \citep{pcaob_as1105}. Legal professional guidance treats generative AI outputs as requiring competent use, confidentiality protection, supervision, and appropriate verification by lawyers \citep{aba2024}. Clinical decision support guidance distinguishes software functions that may be excluded from the definition of a device from software functions that remain subject to device regulation \citep{fda2026}. These regimes make defensibility part of the value of the output: an audit conclusion that cannot be supported by documentation is not partially valuable; it is not valuable at all. A clinical recommendation that cannot be embedded in the patient's record and clinician review is not clinically actionable as an organizational decision. The accountability infrastructure is constitutive of the output's usability.

\subsection{The Theoretical Gap}

The five literatures reviewed above explain important pieces of the agentic AI boundary problem: digital innovation explains modular recombination, transaction cost economics explains falling market-assembly costs, complementary-assets theory explains value capture under weak appropriability, platform governance explains orchestrator power, and IS control explains how information systems enact organizational authority. None fully explains why some AI-enabled tasks remain integrated when their technical interfaces become modular. The missing construct is accountability. Some tasks can be routed through markets because correctness is cheap to verify and responsibility can transfer. Others remain integrated because defensibility, reviewability, signoff, and responsibility are embedded in cospecialized assets that cannot be cheaply recreated at the interface. The next sections develop this argument.

\section{AGENTIC ORCHESTRATION AS A BOUNDARY-SHIFTING CHANGE}

\subsection{AI Orchestration and Falling Interface-Assembly Costs}

An agentic orchestrator is a platform-mediated AI system that interprets user intent, coordinates external capabilities, and controls routing among complements. Unlike a conventional software interface that waits for a user to select a function, an agentic orchestrator can translate an objective into a sequence of tool calls, select among complements, monitor intermediate outputs, and adapt the workflow. This makes it a boundary-shifting technology because it substitutes for some of the coordination historically performed inside vertical applications. The relevant reduction is not merely technical latency or API convenience but a reduction in the economic costs of interface-mediated market assembly: search, tool invocation, workflow sequencing, schema translation, monitoring, and adaptation all become cheaper, making component markets more viable for tasks that were previously bundled. This dynamic is especially important in knowledge-intensive vertical markets, including accounting, law, healthcare, procurement, insurance, and software engineering, that have historically relied on applications bundling data access, task execution, workflow, permissions, review, and records. Customers may become less willing to pay for workflow assembly if the orchestrator can supply it, but they may remain willing to pay for domain-specific accountability assets that the orchestrator cannot supply.

Two properties of agentic orchestrators interact with the accountability boundary in ways that distinguish them from prior generations of digital platform mediation. First, they exercise \textit{autonomy}: rather than executing predefined workflows, they interpret user intent and select among possible tool-call sequences, including which complements to invoke, in what order, and how to interpret intermediate results. Second, they are \textit{stochastic}: the same input can produce different sequences of tool calls and different intermediate outputs across runs. Together these properties amplify the verification challenge developed below. An output produced by a stochastic, autonomous orchestrator is not simply correct or incorrect but correct or incorrect on this run, with this prompt, given this sequence of tool calls. Verification therefore shifts from output checking to output checking under stochasticity, and accountability shifts from identifying a responsible party to identifying responsibility given the run-specific path that produced the output. The accountability assets introduced in §4 must therefore capture not only evidence of the output but evidence of the path, the tool calls invoked, and the autonomous choices the orchestrator made along the way.

\subsection{The Strategic Response Among Vertical AI Firms}

The strategic response emerging among vertical AI firms is to expose specialized functions as callable capabilities while allowing general-purpose agents to own more of the user-facing workflow. In the language of practitioners, the firm goes ``headless.'' The strategy can be rational when the firm sells a capability whose output is easy to verify, responsibility is transferable, and the orchestrator supplies distribution the firm cannot match alone. In those cases, the interface may not be the source of durable value. The same strategy can be destructive when the firm sells accountable judgment. If the orchestrator controls the evidence trail, review workflow, signoff surface, or responsibility record, moving the interface can separate formal accountability from control over the assets needed to justify, review, and discharge it. A vertical firm can believe it has ceded only the screen while retaining the capability. In practice, the orchestrator may come to control the user relationship, task framing, evidence capture, exception handling, and final signoff surface. The firm must therefore design modular access to preserve control over evidence, review, lineage, signoff, and responsibility. The apparent interface decision becomes a boundary decision.

\subsection{Why Open Protocols Do Not Remove Orchestrator Power}

Open protocols and standardized tool interfaces are important because they make complements easier to expose. They are not sufficient to protect complementors from orchestrator power. A protocol can specify how a tool is described, called, and returned, but it usually does not determine which tool the orchestrator selects, how often it is invoked, how it is ranked, how results are interpreted, or when first-party functionality substitutes for it. The interface is standardized, but the demand allocation mechanism remains controlled by the orchestrator. This pattern parallels findings in earlier digital platform settings, where openness in technical interfaces coexisted with concentrated control over discovery and matching \citep{boudreau2010,tiwana2010,zhu2018}. Specific protocol implementations, such as the Model Context Protocol and the ChatGPT applications interface, illustrate the interface shift without determining the accountability boundary \citep{mcp2025,openai_appssdk}.

The boundary implication is that firms should distinguish between making capabilities callable and ceding control over accountability-relevant assets. Callable interfaces can expand markets. Allowing evidence, review, signoff, lineage, or responsibility records to be created or controlled outside the firm's governed system can erode appropriability. Agentic orchestration therefore creates pressure toward disaggregation, but that pressure is filtered through accountability assets, appropriability, and orchestrator dependence.

\section{ACCOUNTABILITY ASSETS AS LIMITS TO DISAGGREGATION}

\subsection{Definitions and Construct Scope}

This paper introduces accountability assets to explain why some AI-enabled capabilities resist disaggregation. Accountability assets are complementary assets whose strategic value derives from making AI-supported outputs legitimate, auditable, reviewable, and assignable to a responsible party after the fact. Accountability assets often become cospecialized with the AI-enabled capability they support, and the degree of cospecialization varies; that variation is what Proposition 4 mobilizes. Examples include evidence trails, review workflows, professional signoff, escalation procedures, permissions, authoritative repositories, evaluation records, and governed decision rules. These assets are not merely internal controls in a narrow compliance sense. They are value-capture assets because they determine whether the output can be used, defended, and relied upon by the customer organization. We use the term \textit{assets} in a broad IS and strategy sense to include durable socio-technical resources, records, routines, rights, and infrastructures that require investment, can be accumulated or degraded over time, and affect value capture or accountable use. An accountability asset may therefore be a technical artifact, such as an evidence log, or an organizationally enacted resource, such as a signoff routine or escalation right.

Accountability assets matter because AI-supported outputs often have two values. The first is task value: the immediate usefulness of the generated classification, extraction, recommendation, draft, or analysis. The second is accountable-use value: the ability to use that output in an organizational setting where someone must justify the decision. A general-purpose model may approximate task value. It may not approximate accountable-use value if the output lacks evidence, provenance, review, and responsibility. The more the customer buys accountable-use value rather than task completion alone, the more accountability assets shape the boundary.

In IS terms, accountability assets are socio-technical rather than purely economic. They include features of the IT artifact, such as logs, permission structures, lineage records, workflow states, and repositories, together with organizational arrangements, such as review routines, escalation roles, signoff rights, and exception ownership. This framing follows the IS tradition of theorizing the IT artifact in relation to its organizational context \citep{orlikowski2001,strong2010}. It also clarifies why the construct belongs in IS theory: accountability assets are the designed and enacted information systems infrastructure through which accountable organizational use becomes possible.

Table 1 defines the core constructs used in the theory. The definitions keep the analysis at the capability level rather than treating the firm label as the unit of explanation. This is crucial because a vertical firm may expose one capability as a component, retain another inside an integrated accountability boundary, and adopt a dual-track strategy when edge functions and the accountable core differ in their task-accountability regimes.

\begin{table}[ht!]
\centering
\caption{Construct Definitions}\label{tab:construct-definitions}
\renewcommand{\baselinestretch}{1.0}\footnotesize
\begin{tabularx}{\textwidth}{p{1.8in} X}
\toprule
\textbf{Construct} & \textbf{Definition} \\
\midrule
AI-enabled capability & A discrete work activity in which AI substantially performs or supports execution, subject to boundary strategy choices. \\
\addlinespace
Agentic orchestrator & A platform-mediated AI system that interprets user intent, coordinates external capabilities, and controls routing among complements. \\
\addlinespace
Interface boundary & The location of user interaction, task initiation, and workflow routing. \\
\addlinespace
Accountability boundary & The location of responsibility, evidence, review, signoff, and post hoc defensibility. \\
\addlinespace
Accountability assets & Complementary assets whose strategic value derives from making AI-supported outputs legitimate, auditable, reviewable, and assignable to a responsible party after the fact. \\
\addlinespace
Responsibility transferability & The degree to which responsibility for an output can be contractually, professionally, or operationally shifted across organizational boundaries. \\
\addlinespace
Verification cost & The cost to determine whether an AI-supported output is correct, complete, defensible, and usable in accountable organizational action. \\
\addlinespace
Rule debt & The latent governance burden created when organizational decision rules migrate from formal information systems into ungoverned agentic execution environments (such as informal natural-language prompts or agentic workflows), producing unmanaged policy artifacts that must eventually be inventoried, versioned, tested, and reconciled. \\
\bottomrule
\end{tabularx}

\vspace{0.5em}
{\scriptsize\textit{Note. The table defines the newly introduced and most central constructs; additional constructs used in the propositions and operationalizations (e.g., interface-and-assembly cost reduction, appropriability, orchestrator dependence, boundary strategy, value capture, boundary misconfiguration) are defined in the surrounding prose. Definitions are stated at the level of the AI-enabled capability.}\par}
\renewcommand{\baselinestretch}{2.0}\normalsize
\end{table}

\subsection{How Accountability Assets Differ from Adjacent Constructs}

Accountability assets are related to, but distinct from, several adjacent constructs in the IS and strategy literatures. They are a subset of complementary assets because they help turn an AI capability into a usable offering. They often become cospecialized with the AI-enabled capability because their value frequently depends on that capability, and the capability depends on them for legitimate use, but accountability assets need not be cospecialized in every case; the degree of cospecialization varies and is what Proposition 4 mobilizes. They are also not simply a new label for cospecialized complementary assets: their strategic value derives specifically from making outputs defensible, reviewable, and assignable to a responsible party after the fact, properties that are not central to the definition of cospecialized assets generally.

Accountability assets differ from IS control mechanisms. IS control research has identified portfolios of formal and informal controls through which organizations align IS-related work with organizational objectives \citep{kirsch1996,kirsch1997,wiener2016}. Control mechanisms regulate behavior. Accountability assets are resources that enable accountable use of outputs after the fact. Controls and accountability assets often coexist: a workflow may be governed through behavior controls, while the resulting documentation, evidence, and signoff records constitute accountability assets. The constructs operate at different levels. Controls govern how work is done. Accountability assets are the residual organizational resources that make the output of that work defensible.

Accountability assets differ from IT governance arrangements. IT governance theory specifies how decision rights over IT-enabled activities are allocated across organizational actors \citep{sambamurthy1999,weill2004,tiwana2014}. Governance arrangements determine who decides; accountability assets determine whether decisions can be defended. The two are related but distinct. An organization can have well-allocated decision rights and still lack the evidence, review, and signoff infrastructure that makes specific AI-supported decisions accountable.

Accountability assets are also distinct from several adjacent constructs in the strategy and organizations literatures. Relational assets derive value from repeated exchange and trust between specific parties; accountability assets may be relational but need not be, since documentation, signoff, and audit trails create value through their support for review by regulators, courts, or governance bodies regardless of dyadic relationships. Reputational assets signal trustworthiness before exchange; accountability assets produce the evidence and responsibility structure that makes specific outputs defensible after exchange. Compliance capabilities are broad organizational capacities to meet regulatory requirements; accountability assets focus on output-level evidence, review, and responsibility structure for specific AI-enabled capabilities. Dynamic capabilities concern higher-order processes for sensing, seizing, and reconfiguring resources; accountability assets are first-order complementary resources, often cospecialized with the focal capability, that make particular outputs defensible. A firm may use dynamic capabilities to build accountability assets, but the constructs operate at different levels.

\subsection{Cospecialization and the Investment-Incentive Logic}

Accountability assets are often cospecialized because the AI-enabled capability and the accountability asset become mutually dependent. An audit assistant is more valuable when tied to evidence standards, materiality judgments, review history, and signoff workflows; the review infrastructure is more valuable when AI makes evidence assembly, anomaly detection, and documentation more efficient. A clinical decision support capability is more valuable when connected to clinician review, patient context, and escalation protocols; the review infrastructure is more valuable when AI improves triage and recommendations. Neither side is generic once the capability is embedded in accountable work.

This cospecialization changes investment incentives. \citet{williamson1985} emphasized that relationship-specific investments create governance hazards when returns depend on another party, and property-rights theory formalized the intuition that residual control affects investment incentives when contracts are incomplete \citep{grossman1986,hart1990}. Applied here, the party that controls the accountability asset will invest more in improving it if it can appropriate the returns. Orchestrators rationally invest in general planning, routing, and interface design because those investments are reusable across domains. Vertical firms rationally invest in domain-specific evidence, evaluation, signoff, and governed workflows. Customer organizations rationally invest in internal policy, responsibility allocation, and authoritative repositories. Boundary strategy should align residual control with the party whose investment is most necessary for accountable use. When that alignment fails, accountability assets deteriorate or rule debt accumulates.

\subsection{Appropriability and Accountability Assets}

Appropriability moderates the value of accountability assets. When the AI capability is hard to imitate because of proprietary data, tacit expertise, accumulated evaluations, or learning effects, the capability owner can capture value through the capability itself. When appropriability is weak, control over accountability assets becomes a stronger predictor of value capture. This is common in tasks that use frontier models to perform extraction, summarization, classification, or drafting. Competitors and orchestrators can approximate the task output, but may not be able to reproduce the evidence chain, signoff authority, domain-specific review process, or customer trust needed to use the output in accountable work.

This logic does not imply that every computational step must be performed inside the vertical firm. It implies that the accountability boundary should not be surrendered casually. Integrated strategies may rely on external computational services or bounded task execution, but only when accountability-relevant context, evidence, review, lineage, signoff, and responsibility remain controlled by the party that must defend the output. A customer may call external capabilities while retaining responsibility in governed systems. A third party may certify outputs or provide warranties that make responsibility more transferable. The key is that the party best positioned to improve and defend the accountability asset must retain sufficient control to invest in it and appropriate returns from it.

\section{A THEORY OF BOUNDARY STRATEGY UNDER AGENTIC ORCHESTRATION}\label{sec:theory-boundary-strategy}

\subsection{The Two Primary Dimensions}

The theory organizes boundary choice around two primary dimensions: verification cost and responsibility transferability. Verification cost is the cost to determine whether an AI-supported output is correct, complete, defensible, and usable in accountable organizational action. Low verification cost means errors can be detected cheaply, often by inspection, comparison to source material, deterministic validation, or statistical sampling. High verification cost means evaluation requires domain expertise, context, professional judgment, causal assessment, or a review trail that is itself costly to assemble. Although verification can be treated as part of enforcement cost in the broad transaction-cost tradition, we distinguish interface-and-assembly cost reduction from verification cost because agentic orchestrators reduce the former more directly than the latter.

Responsibility transferability is the degree to which responsibility for an output can be contractually, professionally, or operationally shifted across organizational boundaries. High transferability means a customer can rely on warranties, certification, contractual allocation, insurance, or market norms to shift responsibility. Low transferability means legal, professional, ethical, or operational responsibility remains with the customer organization, licensed professional, or integrated provider even if task execution is externalized. Verification cost concerns whether the output can be evaluated. Responsibility transferability concerns whether the obligation to stand behind the output can move.

Table 2 summarizes the roles of the theoretical frameworks in the argument, and Table 3 maps the two dimensions to expected boundary strategies. Verification cost and responsibility transferability are sufficient boundary conditions for integration: either high verification cost or low responsibility transferability shifts the prediction toward retention of the accountability boundary, while P2 and P3 isolate their individually testable effects. Accountability-asset cospecialization adds mechanism rather than a separate primary dimension by explaining why verification becomes costly, responsibility difficult to transfer, and retained control strategically valuable. Appropriability and orchestrator dependence are moderators rather than primary boundary-strategy dimensions. Appropriability affects whether accountability assets become the dominant source of value capture. Orchestrator dependence affects how much value the platform actor can appropriate from the complementor after adoption of a component strategy.

The theory is a capability-level variance theory, not a deterministic sector taxonomy. It assumes that the orchestrator is technically capable of invoking the external capability, that the customer organization cares about using outputs in organizational action rather than merely consuming suggestions, and that accountability for at least some outputs is institutionally meaningful. Under these assumptions, verification cost and responsibility transferability determine whether the execution boundary and accountability boundary can move together. If accountability is institutionally light, or if outputs are purely exploratory and create no organizational commitment, the accountability-assets mechanism should have limited explanatory power.

\begin{table}[ht!]
\centering
\caption{Theoretical Frameworks and Their Roles in the Argument}
\renewcommand{\baselinestretch}{1.0}\footnotesize
\begin{tabularx}{\textwidth}{p{2.2in} X}
\toprule
\textbf{Framework} & \textbf{Role in the paper} \\
\midrule
Digital innovation and modularization & Baseline prediction: modular digital architectures, intensified by agentic orchestration, exert pressure toward organizational disaggregation. \\
\addlinespace
Coase, Williamson, and transaction costs & Mechanism for disaggregation pressure: agentic orchestration lowers interface and assembly costs, while verification costs for accountable use remain a distinct boundary condition. \\
\addlinespace
Teece and complementary assets & Core mechanism: value capture depends on cospecialization with accountability assets when appropriability is weak. \\
\addlinespace
Digital platform governance and envelopment & Strategic hazard: agentic orchestrators can appropriate value through interface control, routing, ranking, and substitution, even under open protocols. \\
\addlinespace
Property-rights theory (Grossman-Hart-Moore) & Investment logic: residual control over accountability assets shapes incentives to invest in them. \\
\addlinespace
IS control and IT governance & Substrate for rule debt: control configurations historically absorbed by governed information systems fragment when agentic AI executes rules outside those systems. \\
\bottomrule
\end{tabularx}

\vspace{0.5em}
{\scriptsize\textit{Note. The table identifies functional roles rather than co-equal centers of gravity.}\par}
\renewcommand{\baselinestretch}{2.0}\normalsize
\end{table}

\begin{table}[ht!]
	\centering
	\caption{Boundary Strategies Under Agentic AI Orchestration}\label{tab:boundary-strategies}
	\renewcommand{\baselinestretch}{1.0}\footnotesize
	\begin{tabularx}{\textwidth}{p{1.75in} p{1.55in} X}
		\toprule
		\textbf{Task-accountability regime} & \textbf{Expected boundary strategy} & \textbf{Strategic logic} \\
		\midrule
		Low effective verification cost; responsibility transferable & Component & The capability or sub-capability can be sold or invoked as a callable component because outputs are cheap to check, or assurance mechanisms make checking less costly for the customer, and responsibility can move through routine market, contractual, or assurance mechanisms. \\
		\addlinespace
		High effective verification cost and/or responsibility not transferable & Integrated & The accountability boundary is retained because accountable use depends on evidence, review, lineage, signoff, authoritative records, and assignable responsibility that cannot be reconstructed by an orchestrator at runtime. External computational services or modular front doors may be used only within controlled interfaces that preserve accountability-relevant traces. \\
		\addlinespace
		Mixed capability bundle: componentizable edge functions plus accountability-bearing core & Dual-track & Separable edge functions are exposed as components while the accountable core remains integrated. The component track optimizes for adoption and interoperability; the integrated track optimizes for defensibility and responsibility. \\
		\bottomrule
	\end{tabularx}
	
	\vspace{0.5em}
	{\scriptsize\textit{Note. The table identifies three boundary strategies. Bounded external computation inside a retained accountability boundary is an implementation of integration, not a separate top-level category.}\par}
	\renewcommand{\baselinestretch}{2.0}\normalsize
\end{table}

\subsection{Component Strategy}\label{sec:component-strategy}

An AI-enabled capability is likely to adopt a component strategy when effective verification cost is low and responsibility is transferable. In this strategy, the orchestrator can call the capability as a market component, and the customer can cheaply evaluate the output or shift responsibility through contract or routine operational processes. Examples include data extraction from standardized documents, document classification, first-pass retrieval, simple formatting, and low-stakes summarization. The component provider competes on accuracy, reliability, availability, latency, price, security, and ease of integration. A component strategy does not imply strategic irrelevance: a component provider can capture value through accuracy advantages, proprietary feedback loops, privileged data access, trust, or scale across orchestrators. The strategic challenge is commoditization, not loss of accountability assets. A component strategy can also occur when raw verification cost is high but certification, warranties, insurance, or third-party assurance lower the customer's effective verification burden and make responsibility transferable. The boundary strategy is then a component strategy with assurance.

\subsection{Integrated Strategy}\label{sec:integrated-strategy}

Integration is more likely when effective verification cost is high or responsibility is not transferable. In this strategy, the customer cannot cheaply determine whether the AI-supported output is correct, or cannot shift responsibility to the orchestrator or external component provider even when the output can be checked. The capability must remain tied to evidence, review, signoff, lineage, permissions, and authoritative records. Integration is justified not because the interface is technically difficult, but because accountable use requires complementary assets whose value depends on the capability and whose control affects investment incentives. Audit evidence and signoff illustrate the pattern: a general agent may initiate an audit request or call bounded analyses, but the audit conclusion depends on evidence quality, materiality, review, documentation, and professional responsibility, assets that cannot be assembled at runtime by a general orchestrator without the accountable infrastructure that gives the conclusion standing. Similar logic applies to clinical decision support, high-stakes safety decisions, and regulated approvals. Integrated strategies may still rely on external computational services, bounded task execution, or modular front doors, but only inside controlled interfaces that preserve accountability-relevant context, evidence, logs, review, lineage, signoff, and responsibility under the vertical firm, customer organization, or accountable professional context. The agent becomes a channel, request surface, or workflow assistant rather than the locus of responsibility.

\subsection{Dual-Track Strategy}\label{sec:dual-track-strategy}

Many AI-enabled capability bundles face a mixed task-accountability regime rather than a single boundary choice. A dual-track strategy arises when separable edge functions have low effective verification costs and transferable responsibility, while the accountability-bearing core has high verification costs, non-transferable responsibility, or cospecialized accountability assets. The split is therefore not between external execution and internal signoff for the same accountability-bearing output. The split is between sub-capabilities in the offering portfolio. Edge functions such as extraction, retrieval, citation checking, document comparison, scoring, and first-pass drafting can be exposed as components. Core commitments such as audit signoff, legal opinions, clinical care planning, supplier approval, and fiduciary recommendations remain integrated because their value depends on evidence, review, lineage, signoff, and assignable responsibility.

Dual-track strategy clarifies why the same vertical AI firm can both make some capabilities headless and retain integration around others. The firm may make document intelligence or other edge functions easy to call while retaining the audit platform, legal work product, clinical workflow, or procurement approval system that makes final commitments defensible. This is a capability-bundle strategy that decomposes by accountability rather than by interface. The integrated core may be reachable through bounded modular interfaces for initiation, input submission, status retrieval, or delivery, but those interfaces do not transfer control over evidence, review, lineage, signoff, system-of-record updates, or final commitment. External computation remains an implementation of integration only if accountability-relevant traces and decision rights stay inside the retained accountability boundary; otherwise the firm has surrendered the value-capture surface the theory identifies as consequential.

\subsection{Rule Debt as a Governance-Cost Consequence}

Rule debt is the latent governance burden created when organizational decision rules migrate from formal information systems into ungoverned agentic execution environments, such as informal natural-language prompts or agentic workflows, producing unmanaged policy artifacts that must eventually be inventoried, versioned, tested, and reconciled. The construct intentionally builds on two threads in the IS literature. First, the metaphor of technical debt captures the future rework that expedient software choices create \citep{cunningham1992}, and research on hidden technical debt in machine learning systems has shown that entanglement, data dependencies, configuration, and monitoring problems accumulate in ML systems specifically \citep{sculley2015}. Second, the IS control literature has analyzed how control configurations are enacted through information systems and how those enactments shape organizational outcomes \citep{kirsch1996,kirsch1997,wiener2016}. Rule debt extends both threads by identifying a specific form of governance debt that arises when control-bearing rules execute outside the information systems that previously enacted them. Rule debt is the direct consequence of what we term \textit{boundary misconfiguration}: the organizational error of adopting a component strategy for a capability that, due to low responsibility transferability or high verification cost, requires retained control over the accountability boundary. When a firm or customer aggressively modularizes accountability-bearing work without preserving governed evidence, review, lineage, signoff, and rule ownership, the resulting gap is filled by ungoverned rule debt.

A prompt or agent instruction can encode a business rule by specifying when to escalate a claim, classify a transaction, summarize a medical message, rank a supplier, approve an exception, or draft a legal response. When such instructions are created informally, they become policy fragments outside the customer organization's governed information systems. At small scale, this appears flexible. At organizational scale, it creates an ungoverned control environment in which rules must be found, owned, versioned, tested, reconciled with formal policy, monitored for drift, and connected to accountability. Otherwise, the customer accumulates governance costs that were previously absorbed by integrated systems. Rule debt is most likely when the boundary separates decision-rule execution from governed accountability assets: if an orchestrator executes rules while the customer retains responsibility, and if neither the orchestrator nor the vertical firm owns rule governance, the customer inherits the burden. The task boundary moved to a market component, but the accountability boundary did not move with it.

The construct has direct implications for IS control theory, which the paper develops further in §9. Wiener et al.\ (2016) emphasized that control configuration and control enactment are distinct phenomena. Rule debt is, in part, the cost of allowing the enactment of control to migrate outside the configuration the organization has formally adopted.

\subsection{Boundary Migration Over Time}\label{sec:scope-boundary-conditions}

Boundary strategies are dynamic. Capabilities can migrate from integrated to component strategies when outputs standardize, verification becomes cheaper, regulators accept new control frameworks, third-party assurance develops, and customers accept new allocations of responsibility. Firms can also migrate from integrated to dual-track strategies when edge functions become componentizable before the accountability-bearing core does. Conversely, capabilities can migrate toward integration when failures reveal hidden verification costs, liability becomes salient, or rule debt accumulates. The theory predicts boundary strategies conditional on the state of verification, responsibility transferability, appropriability, and orchestrator dependence, not permanent classifications by sector.

Migration is shaped by the development of institutional infrastructure. As assurance institutions mature, including third-party certification, professional warranties, insurance markets, regulatory safe harbors, and industry standards, responsibility becomes more transferable for tasks previously bound to integrated providers. As verification technologies improve, verification costs fall. Both forces push capabilities toward components. The robo-advisory case illustrates institutional accommodation rather than technological substitution alone. Algorithmic advisory services operated within registered-investment-adviser frameworks, while SEC staff guidance clarified expectations around disclosure, collection of client information, algorithmic oversight, and compliance programs \citep{sec2017}. The trajectory supports the theory's migration logic: the component strategy becomes more feasible when institutions preserve accountability while allowing execution to become more automated. Migration also runs in the opposite direction: capabilities that appeared safely componentizable can migrate back toward integration when failures reveal hidden verification costs, when regulators require human review, or when customers insist on integrated responsibility.

A further migration pattern is generated by the rule debt mechanism. As customer organizations accumulate rule debt from component strategies, the implicit cost of those strategies rises, producing a delayed migration toward integration as customers come to value the absorbed governance services that integrated providers had previously supplied. The timing asymmetry yields two temporal implications. First, when verification technologies and assurance institutions improve, migration should begin at the execution layer before it reaches the commitment layer: components become callable before responsibility fully transfers. Second, when failures reveal hidden verification costs or when rule debt becomes visible, migration should reverse at the accountability layer before it reverses at the interface: customers demand governed evidence, review, and signoff even if they continue to use external components for task execution. These movements are path dependent because investments in evidence records, exception histories, and policy repositories accumulate over time.

\section{PROPOSITIONS}

The propositions link the causal chain from agentic reductions in interface and assembly costs to component-strategy pressure, then to accountability-based boundary conditions, value capture, complementor dependence, and customer-side governance costs. They are stated at the level of the AI-enabled capability. Each proposition is derived from the prior literature, specifies a mechanism, identifies the alternative prediction it modifies, and notes scope conditions.

\subsection{Interface-Assembly Cost Reduction and the Component Strategy}

\textbf{Proposition 1.} \textit{The greater the reduction in interface and assembly costs produced by agentic orchestrators, the more likely AI-enabled capabilities are to adopt a component strategy, all else equal.}

P1 follows directly from transaction cost logic and from digital innovation theory's modularization predictions. \citet{coase1937} argued that firms exist because using markets is costly, and \citet{williamson1985} identified the specific hazards, including bounded rationality, opportunism, asset specificity, and uncertainty, that make integration attractive. Digital innovation theory leads us to expect that modular architectures and recombinant logics reduce the costs of decentralized assembly \citep{yoo2010,henfridsson2013}. When the interface and assembly costs that historically justified integration decline, market governance becomes relatively more attractive. Agentic orchestrators directly attack several cost categories the literature treats as central: search, tool invocation, sequencing, schema translation, monitoring, and adaptation all become cheaper. The mechanism is reduction in the costs of market assembly for tasks that were previously bundled inside vertical applications.

The alternative prediction this proposition modifies is the assumption that AI capability advances translate into capability concentration. A countervailing prediction is that frontier models will absorb vertical functionality, producing integration around general-purpose agents. The prediction grounded in transaction cost economics and digital innovation theory is the opposite: by lowering the cost of using markets, orchestration favors specialization where specialization is technically possible. The scope condition is reliability. The effect should be strongest where orchestrators have reached operational reliability for the task class and weakest where orchestration is technically possible but operationally unreliable.

\subsection{Verification Cost as a Boundary Condition}

\textbf{Proposition 2.} \textit{The positive relationship between agentic reductions in interface and assembly costs and adoption of a component strategy weakens as verification costs increase.}

P2 introduces the first accountability boundary condition. Lower interface and assembly costs make market assembly easier, but they do not determine whether the assembled output is usable in accountable organizational action. In the broad transaction-cost tradition, verification can be treated as part of enforcement cost; here it is separated because agents often reduce the cost of calling and composing capabilities without equivalently reducing the cost of defending their outputs. When evaluation requires domain expertise, context, professional judgment, causal assessment, or assembled evidence, the customer cannot cheaply confirm that the work product meets the standard required for organizational use. An output that cannot be verified cheaply is not interchangeable with one produced inside an integrated provider, even when the technical interface is identical.

This proposition modifies an assembly-cost-only prediction that would treat search, interface assembly, and ordinary monitoring costs as the primary boundary determinants. It extends, rather than contradicts, transaction cost theory. In knowledge-intensive AI-supported work, correctness and defensibility may not be visible without expert evaluation, even after use. If verification difficulty is high only because no one has invested in evaluation tooling, the effect should weaken as evaluation technology improves. Where the difficulty stems from the nature of the judgment, such as clinical, legal, audit, or fiduciary judgment, the effect should be persistent.

\subsection{Responsibility Transferability as a Boundary Condition}

\textbf{Proposition 3.} \textit{The positive relationship between agentic reductions in interface and assembly costs and adoption of a component strategy weakens as responsibility for outputs becomes less transferable across organizational boundaries.}

P3 introduces the second accountability boundary condition. Even when an output can be verified, the obligation to stand behind it may remain with the customer, professional, or integrated provider. Responsibility transferability is conceptually distinct from verification cost. A customer can verify an output and still be unable to transfer responsibility for it. This typically arises when professional, regulatory, or fiduciary obligations attach to the actor rather than to the task. A lawyer remains responsible for a filed brief even if the citation retrieval was perfectly verified. A physician remains responsible for a clinical decision even if the underlying recommendation was generated by an external system. The mechanism is that externalization saves interface and assembly cost but does not redistribute legal, professional, or operational accountability, so the customer retains the residual exposure that integration historically absorbed.

This proposition modifies a strict reading of platform openness theory, which sometimes implies that open interfaces and contractual instruments can fully reallocate exposure. In settings of low responsibility transferability, contractual reallocation is incomplete because responsibility is institutionally anchored to particular actors and roles. The scope condition concerns the institutional environment. Where assurance institutions, certification regimes, or insurance markets develop to absorb responsibility, transferability rises and the effect weakens. Where responsibility remains tied to professional, regulatory, or fiduciary roles that cannot delegate substantive judgment, the effect is durable. This proposition predicts persistent integration not because firms cannot decompose the task technically but because they cannot decompose the responsibility institutionally.

\subsection{Accountability-Asset Cospecialization and Retention of the Accountability Boundary}

\textbf{Proposition 4.} \textit{The greater the cospecialization between an AI-enabled capability and accountability assets, the more likely the relevant organization is to retain control over the accountability boundary; when a capability bundle contains componentizable edge functions and an accountability-bearing core, this retention is more likely to take a dual-track form.}

P4 states the core complementary-assets claim of the paper. \citet{teece1986} argued that when an innovation is imitable, value flows to controllers of complementary assets. Accountability assets serve a distinct function: making outputs defensible after use rather than getting them to market. When an AI-enabled capability and its accountability assets are mutually dependent, neither is fully usable without the other. An audit AI without governed evidence is not a useful audit AI; a clinical AI without clinician review infrastructure is not a useful clinical AI. The mechanism combines Teece's appropriability logic with the property-rights insight that residual control should attach to the party whose investment most affects asset quality \citep{grossman1986,hart1990}. If a vertical firm, customer organization, or accountable professional loses control over accountability assets, it loses both value capture and the incentive to improve them.

The proposition is deliberately not stated as ``the firm remains fully internalized.'' The claim is sharper: the relevant organization retains control over the accountability boundary. This control can be exercised through fully internal execution or through bounded external computation inside a controlled accountability envelope. In both cases, accountability-relevant context, evidence, review, lineage, signoff, and responsibility remain governed by the party that must defend the output. When separable edge functions have low verification cost and transferable responsibility, the predicted capability-bundle strategy is dual-track: edge functions can be exposed as components while the accountability-bearing core remains integrated. The alternative prediction this proposition modifies is the inference that technical decomposability implies organizational disaggregation. The scope condition is genuine cospecialization: where accountability assets require direct connection to the capability's outputs, evaluations, and exception cases, the effect should be strong; where this dependence is only a contingent design choice, the effect should weaken.

\subsection{Appropriability and Accountability-Asset Value Capture}\label{sec:appropriability}

\textbf{Proposition 5.} \textit{When appropriability of the AI capability is weak, control over accountability assets becomes a stronger predictor of value capture.}

P5 follows from Teece's appropriability logic but extends it to AI-enabled capabilities. When competitors and orchestrators can approximate task output, value capture cannot rest on capability ownership alone. The strategic question becomes which cospecialized assets can sustain returns under imitation. Accountability assets typically meet this criterion: they accumulate over time through use, embed customer-specific evidence and review records, are anchored in professional or regulatory regimes that are slow to imitate, and acquire value from ongoing operational integration that is costly to replicate. The mechanism is that imitation pressure on the AI capability shifts the locus of appropriable rents to cospecialized assets that are more resistant to imitation. The competing prediction this proposition modifies is that scale, data, or model quality alone are sufficient defenses against frontier model commoditization. Scale and data advantages can be durable but are also subject to erosion as frontier models improve. Accountability assets provide a complementary defense whose value mechanism is different: they protect value capture not by making the model better but by making the output usable in accountable settings. The scope condition is the appropriability regime of the focal capability. Where the AI capability itself enjoys strong appropriability, through proprietary data, regulatory exclusivity, network effects, or substantial switching costs, the differential value of accountability assets is smaller. The proposition is comparative across capabilities within the same firm: empirically, accountability-asset dominance should appear as stronger retention, pricing power, and customer willingness to pay for capabilities whose evidence, review, and signoff infrastructure is difficult to reproduce, even when raw task performance converges across vendors.

\subsection{Orchestrator Dependence and Intent Capture}\label{sec:orchestrator-intent-capture}

\textbf{Proposition 6.} \textit{The greater the share of user goal formulation, invocation history, and feedback observed by an agentic orchestrator rather than by the vertical complementor, the greater the orchestrator's ability to appropriate value from the complementor through reranking, substitution, or envelopment.}

Here, user intent refers to the task objective and contextual demand signal revealed through prompts, goal refinements, tool selections, feedback, and workflow history, not to an unobservable mental state. P6 links platform envelopment to the unique architecture of agentic systems. The classical holdup logic predicts that relationship-specific investments create appropriable quasi-rents \citep{klein1978,williamson1985}. Agentic orchestrators recreate this hazard not just through demand allocation, but through the capture of user context and intent. Even when the interface is standardized through an open protocol such as the Model Context Protocol \citep{mcp2025}, the orchestrator retains residual control over the user's natural-language requests, goal formulation, and follow-up interactions. A complementor that tunes its capability to one orchestrator does not merely surrender distribution; it surrenders the contextual data required to understand why the user invoked the tool, what alternatives were considered, and which intermediate outputs satisfied the user. The mechanism is that by intermediating user intent, the orchestrator accumulates learning signals that can inform routing, ranking, product design, and, where data-use terms permit, first-party functionality that bypasses the complementor, reducing the complementor to a commoditized execution layer and creating envelopment exposure that proprietary-API lock-in alone could not produce.

The alternative prediction this proposition modifies is a reading of open-protocol literature in which technical openness substantially reduces dependence. Open protocols reduce one form of dependence, the lock-in created by proprietary technical interfaces, but they do not address the dependence created by concentrated intent capture. \citet{boudreau2010} noted the distinction between granting access and devolving control, and \citet{zhu2018} showed empirically how platform actors can use their position to compete with complementors. P6 extends this logic to agentic orchestration by identifying intent capture as a distinct mechanism through which value appropriation proceeds. The scope condition concerns the structure of intent capture. Where complementors retain a direct user relationship sufficient to observe goals, alternatives, and satisfaction, the effect weakens. Where the orchestrator sits between the user and every tool invocation and accumulates intent over time, the effect is strong even under nominal interface openness.

\subsection{Boundary Misconfiguration and Rule Debt}

\textbf{Proposition 7.} \textit{When accountability-bearing capabilities characterized by low responsibility transferability or high verification cost are pushed into component strategies without retained accountability infrastructure or equivalent assurance, the resulting boundary misconfiguration accelerates the accumulation of customer-side rule debt.}

P7 states the secondary contribution. Technical debt research shows that expedient software choices create future rework costs \citep{cunningham1992,sculley2015}; IS control research shows that control enactment can diverge from formally adopted configurations \citep{wiener2016}. Rule debt extends both literatures to organizational decision rules in agentic AI settings. When agent instructions execute rules that previously lived in governed information systems, they become policy fragments outside formal governance infrastructure. They must be inventoried, owned, versioned, tested, reconciled with policy, monitored for drift, and connected to accountability. If they are not, rule execution has moved to a component or orchestrator-controlled environment while the accountability boundary has not, leaving the customer with rules whose governance no longer matches their organizational consequence.

The alternative prediction this proposition modifies is that the costs of adopting a component strategy accrue mainly to the component provider or orchestrator. P7 reallocates these costs: the party that retains responsibility, typically the customer, also retains the governance burden. The scope condition concerns the customer's policy infrastructure. Mature governance for agent instructions limits rule debt; adoption without such governance accelerates it. Providers that supply instruction registries, rule versioning, behavior logs, testing environments, exception ownership, and policy reconciliation can capture value by reducing customer-side rule debt.

\section{THE THEORETICAL MODEL}

Figure 1 presents the theoretical model. Agentic orchestration is the focal technological change that lowers interface and assembly costs and exerts pressure toward the component strategy (P1). Two boundary conditions moderate the relationship: verification cost (P2) and responsibility transferability (P3). Cospecialization with accountability assets predicts retention of the accountability boundary and the possible emergence of dual-track strategy when edge functions differ from the accountable core (P4). Accountability assets shape value capture under weak appropriability (P5). Orchestrator dependence captures the platform-power channel through which value can be appropriated after adoption of a component strategy via intent capture (P6). Boundary misconfiguration generates rule debt as a customer-side governance cost (P7). The model captures both the static prediction, which boundary strategy the theory predicts under given conditions, and the dynamic prediction that capabilities can migrate as institutional infrastructure for verification and responsibility transferability evolves.

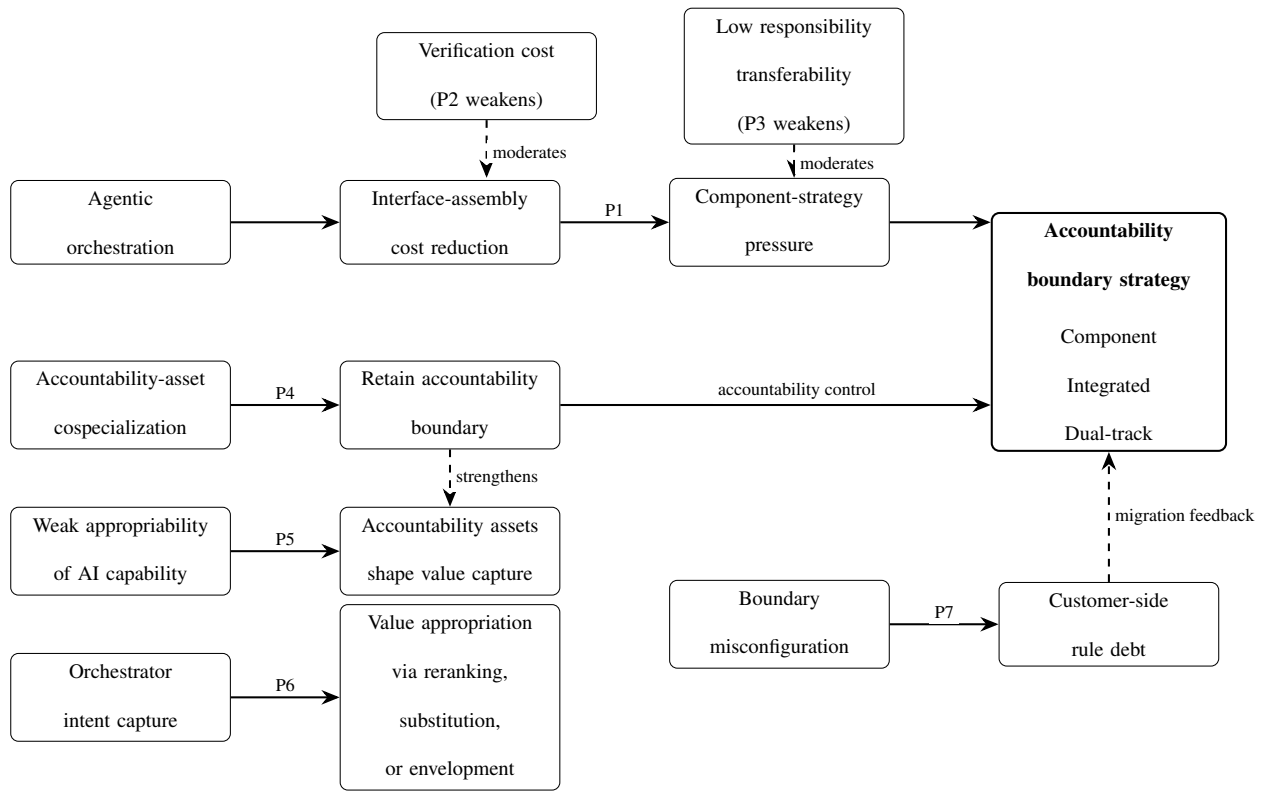
\begin{figure}[ht!]
\centering
\resizebox{\textwidth}{!}{%
\begin{tikzpicture}[
    font=\fontsize{8pt}{9.5pt}\selectfont,
    box/.style={
        draw,
        rounded corners=3pt,
        minimum height=1.1cm,
        minimum width=3cm,
        align=center,
        text width=2.7cm,
        inner sep=4pt
    },
    bigbox/.style={
        draw,
        rounded corners=3pt,
        minimum height=3cm,
        minimum width=3.2cm,
        align=center,
        text width=2.9cm,
        line width=0.8pt
    },
    arr/.style={-{Stealth[length=2.5mm,width=2mm]}, thick},
    darr/.style={-{Stealth[length=2.5mm,width=2mm]}, thick, dashed},
    elabel/.style={font=\fontsize{7pt}{8pt}\selectfont, fill=white, inner sep=2pt}
]
\node[box] (verif) at (5, 6) {Verification cost\\(P2 weakens)};
\node[box] (resp) at (9.2, 6) {Low responsibility\\transferability\\(P3 weakens)};
\node[box] (agentic) at (0, 4) {Agentic\\orchestration};
\node[box] (coord) at (4.5, 4) {Interface-assembly\\cost reduction};
\node[box] (comp) at (9, 4) {Component-strategy\\pressure};
\node[bigbox] (boundary) at (13.5, 2.5) {\textbf{Accountability}\\\textbf{boundary strategy}\\[3pt]Component\\Integrated\\Dual-track};
\node[box] (cospec) at (0, 1.5) {Accountability-asset\\cospecialization};
\node[box] (retain) at (4.5, 1.5) {Retain accountability\\boundary};
\node[box] (approp) at (0, -0.5) {Weak appropriability\\of AI capability};
\node[box] (assets) at (4.5, -0.5) {Accountability assets\\shape value capture};
\node[box] (orchinv) at (0, -2.5) {Orchestrator\\intent capture};
\node[box] (orchval) at (4.5, -2.5) {Value appropriation\\via reranking,\\substitution,\\or envelopment};
\node[box] (miscon) at (9, -1.5) {Boundary\\misconfiguration};
\node[box] (ruledebt) at (13.5, -1.5) {Customer-side\\rule debt};
\draw[arr] (agentic) -- (coord);
\draw[arr] (coord) -- node[elabel, above] {P1} (comp);
\draw[arr] (comp) -- (boundary.west |- comp.east);
\draw[darr] (verif) -- node[elabel, pos=0.55, right] {moderates} (comp.north -| verif.south);
\draw[darr] (resp) -- node[elabel, pos=0.55, right] {moderates} (comp.north -| resp.south);
\draw[arr] (cospec) -- node[elabel, above] {P4} (retain);
\draw[arr] (retain.east) -- node[elabel, sloped, above, pos=0.55] {accountability control} (boundary.west |- retain.east);
\draw[darr] (retain) -- node[elabel, right] {strengthens} (assets);
\draw[arr] (approp) -- node[elabel, above] {P5} (assets);
\draw[arr] (orchinv) -- node[elabel, above] {P6} (orchval);
\draw[arr] (miscon) -- node[elabel, above] {P7} (ruledebt);
\draw[darr] (ruledebt) -- node[elabel, right] {migration feedback} (boundary.south -| ruledebt.north);
\end{tikzpicture}%
}
\caption{Theoretical model of accountability-boundary strategy in agentic AI ecosystems}\label{fig:theoretical-model}
\renewcommand{\baselinestretch}{1.0}\normalsize

\vspace{0.5em}
{\scriptsize\textit{Note. Solid arrows indicate predicted relationships. Dashed arrows indicate moderators or dynamic feedback.}\par}
\renewcommand{\baselinestretch}{2.0}\normalsize
\end{figure}

\section{STRUCTURED THEORETICAL ILLUSTRATIONS}\label{sec:structured-illustrations}

The following illustrations are not empirical tests. They are structured theoretical illustrations, selected to maximize contrast across domains and to discipline the boundary strategy logic. Each illustration is coded on verification cost, responsibility transferability, accountability-asset cospecialization, appropriability, orchestrator dependence, and predicted boundary strategy. For each illustration we describe the mechanism in operation, identify the alternative prediction generated by prior frameworks alone, and explain why the observed pattern tracks the theory. Table 4 presents the coding.

\begin{table}[ht!]
\centering
\caption{Structured Theoretical Illustrations}\label{tab:illustrations}
\renewcommand{\baselinestretch}{1.0}\normalsize
\scriptsize
\setlength{\tabcolsep}{3pt}
\begin{tabular}{p{1.0in} p{0.5in} p{0.85in} p{0.85in} p{0.7in} p{0.7in} p{0.8in}}
\toprule
\textbf{Illustration} & \textbf{Verif.\ cost} & \textbf{Resp.\ transfer.} & \textbf{Account.\ cospec.} & \textbf{Appropri-ability} & \textbf{Orch.\ dependence} & \textbf{Predicted strategy} \\
\midrule
Document extraction & Low & High & Low (unless regulated) & Weak/mod. & High if single agent & Component \\
\addlinespace
Legal research retrieval & Low for retrieval output & High for retrieval output & Moderate (source linkage, provenance) & Moderate (proprietary content) & Mod./high & Component \\
\addlinespace
Audit evidence and signoff & High & Low & High (evidence, materiality) & Weak (AI); stronger (evidence) & Risky if workflow moves & Integrated \\
\addlinespace
Clinical decision support & High & Low & High (patient context, review) & Variable & Risky if outside clinical gov. & Integrated \\
\addlinespace
Legal drafting vs.\ filed opinion & Low--mod.\ (draft); high (commit) & Low (final commit) & High (final stage) & Weak (draft); stronger (review) & Mod. & Dual-track \\
\addlinespace
Procurement scoring vs.\ approval & Mod.\ (score); high (approval) & Low (final approval) & High (policy, exceptions) & Weak/mod. & Mod.\ if platform controls & Dual-track \\
\bottomrule
\end{tabular}
\normalsize

\vspace{0.5em}
{\scriptsize\textit{Note. Illustrations are theoretically sampled comparisons, not empirical tests. Verif.~cost = verification cost; Resp.~transfer.~= responsibility transferability; Account.~cospec.~= accountability-asset cospecialization; Mod.~= moderate; Orch.~dependence = orchestrator dependence and intent capture. The last two rows describe dual-track strategies in which separable edge functions can be sold or invoked as components while the accountable core remains integrated. ``Legal drafting vs.~filed opinion'' contrasts componentizable drafting support against the filed legal commitment. ``Procurement scoring vs.~approval'' contrasts componentizable scoring against governed supplier approval. The predicted strategy column reflects the strategy that integrates both tracks.}\par}
\renewcommand{\baselinestretch}{2.0}\normalsize
\end{table}

\subsection{Document Extraction}

Document extraction is the clearest component-strategy case. The task is to extract fields, classify documents, or convert semi-structured input into a standardized output. Verification cost is relatively low because outputs can be compared to source documents, validated against schemas, or sampled statistically. Responsibility is transferable through service-level commitments, warranties, reprocessing, or customer review. Accountability-asset cospecialization is limited unless extraction directly supports a regulated conclusion. A pure platform-power analysis would predict that orchestrator dependence is the dominant strategic concern. That concern is real, but the more fundamental challenge is commoditization: several orchestrators and frontier models can extract structured fields, and the task does not generate the cospecialized accountability assets that would limit imitation. The predicted boundary strategy is component, and the strategic priority is durable accuracy and integration breadth, not retention of an accountability boundary that the task does not generate.

\subsection{Legal Research Retrieval}

Legal research retrieval is more nuanced because legal use remains professionally governed while retrieval can be self-contained. We code the illustration at the retrieval-capability boundary: the output is retrieved authority or a retrieval summary, not the lawyer's advice, filing, or opinion. Legal guidance requires competent use and independent verification of generative AI outputs \citep{aba2024}, but that downstream activity is a separate accountable capability. For the retrieval output, verification cost is low because users can check authorities, passages, and source coverage; responsibility transferability is high because providers can warrant search coverage, citation accuracy, provenance, and database reliability. An assembly-cost-only reading therefore holds for retrieval: it can be exposed as a component. Professional responsibility constrains downstream use but does not, by itself, shift retrieval into dual-track. If a vendor bundles retrieval with drafting, advice, or filed-work-product commitments, the resulting capability bundle would be dual-track.

\subsection{Audit Evidence and Signoff}

Audit evidence and signoff represent the integrated end of the strategy space. AI can retrieve invoices, compare contracts, flag anomalies, or draft documentation, but the audit conclusion depends on evidence quality, review, materiality, professional standards, and signoff. PCAOB audit documentation \citep{pcaob_as1215} and evidence \citep{pcaob_as1105} standards make the evidence trail central to review and support for conclusions. Verification cost is high, responsibility transferability is low, and accountability assets are highly cospecialized with the AI-enabled capability. Under an assembly-cost-only reading, the interface-and-assembly cost reductions AI delivers in audit work should produce significant pressure toward component strategy. The case is striking precisely because that prediction fails. Despite technical decomposability of evidence retrieval, anomaly detection, and workpaper drafting, the audit conclusion does not modularize. The mechanism is the cospecialization of the audit capability with evidence trails, materiality judgments, review history, and signoff authority; accountability assets that are constitutive of the output's organizational meaning. Without the evidence trail, the conclusion has no standing. The predicted boundary strategy is integration. Bounded computational tasks, such as retrieval, comparison, and anomaly flagging, may be externally supplied or exposed through controlled interfaces, but evidence, review, lineage, signoff, and accountability-relevant traces remain inside governed audit infrastructure. The case illustrates that technical and organizational modularizability are distinct, and that accountability cospecialization is the source of the divergence.

\subsection{Clinical Decision Support}\label{sec:clinical-illustration}

Clinical decision support also illustrates high verification cost and low responsibility transferability. AI may summarize patient histories, suggest differential diagnoses, triage messages, or recommend next steps. Yet clinical use depends on patient context, professional judgment, evidence provenance, escalation, and care responsibility. FDA guidance on clinical decision support underscores that software functions are embedded in regulatory distinctions between device and non-device functions, intended users, and use contexts \citep{fda2026}. A platform-power analysis would predict that orchestrators with strong distribution can envelope clinical decision support. An assembly-cost-only analysis would predict a component strategy as interface and assembly costs fall. Both predictions face the same obstacle: the clinical recommendation is not a transferable unit of output. Its meaning depends on integration with patient records, clinician review, escalation pathways, and care responsibility. The mechanism is cospecialization of the AI capability with clinical accountability infrastructure, combined with low responsibility transferability anchored in professional and regulatory regimes. The predicted strategy is integration. Task support may use modular computational services under clinical control, but final clinical responsibility remains governed. The case generalizes to a broader principle: where outputs are embedded in role-anchored responsibility, even substantial reductions in interface and assembly costs do not produce a component strategy at the level of the responsible decision.

\subsection{Legal Drafting Versus Filed Legal Opinion}

Legal drafting versus a filed legal opinion illustrates dual-track boundary logic. Drafting clauses, comparing documents, and generating first-pass arguments can be componentized because the lawyer can review and edit the work. The filed opinion, executed contract position, or court submission is different: it carries professional responsibility, client-specific judgment, and potential tribunal-facing obligations. An undifferentiated task-level prediction would treat ``legal drafting'' as a single task and ask whether it should be componentized. The dual-track structure shows the limits of that framing. The underlying drafting work generates two distinct outputs: the draft itself, used as an input to lawyer review, and the filed opinion, which is a separate downstream commitment with its own accountability profile. As an input to lawyer review, the draft is componentizable: verification cost is moderate and responsibility remains with the lawyer regardless of how the draft was produced. The filed opinion is a distinct output that, while it may incorporate the draft, carries professional and tribunal-facing accountability that cannot transfer. The two outputs occupy different accountability regimes, and the predicted strategy is therefore dual-track: drafting and comparison can be sold or invoked as component functions, while the filed commitment remains an integrated accountable service.

\subsection{Procurement Risk Scoring Versus Supplier Approval}

Procurement risk scoring versus supplier approval provides a nonprofessional example. A risk-scoring capability can be called as a component to aggregate sanctions data, financial indicators, cyber signals, sustainability metrics, or delivery risk. Supplier approval is a governed organizational decision that affects contracts, compliance, operational resilience, and reputational exposure. A modularity-only prediction would treat supplier evaluation as a single workflow and predict a component strategy as scoring quality improves. The case shows why that prediction is incomplete. The scoring task and the approval task have different verification costs and different responsibility profiles. A score is verifiable by comparing inputs to outputs and is supportable by service-level commitments. An approval is a governed decision whose responsibility rests with named individuals and committees and is embedded in contract policy, exception records, audit trails, and accountable ownership. The mechanism is the same as in the legal case: separable analytical and commitment functions participate in different accountability regimes. The predicted strategy is dual-track: component scoring with governed approval. Practitioner evidence is consistent with the prediction. Reported descriptions of Walmart's AI-driven supplier-negotiation system show automated negotiation handling bounded supplier dialogue while the organization retains the negotiation envelope, escalation conditions, and approval structure \citep{vanhoek2022}. The case extends the theory beyond professional services to organizational decisions in supply chain management.

\section{IMPLICATIONS}

\subsection{Implications for Information Systems Theory}

This paper makes three contributions to information systems theory. First, it qualifies digital innovation theory's expectations about modularization. Digital innovation theory has emphasized that modular architectures and recombinant logics support more decentralized organizing forms \citep{yoo2010,henfridsson2013,lyytinen2016}. The theory developed here specifies a structural condition under which this expectation fails: technical decomposability does not entail organizational decomposability when accountability assets are cospecialized with the focal capability and when responsibility cannot transfer across the technical interface. Accountability assets are a previously undertheorized class of complementary assets that often become cospecialized with the focal capability, whose presence determines whether agentic orchestration produces the modular organizational form its architecture would predict.

Second, the theory extends IS control theory and IT governance theory. IS control research has analyzed how organizations exercise authority over IS-related work through portfolios of formal and informal controls \citep{kirsch1996,kirsch1997,choudhury2003,wiener2016}, and IT governance research has specified how decision rights over IT-enabled activities should be allocated \citep{sambamurthy1999,weill2004,tiwana2014}. Agentic AI complicates both. When decision rules execute through agents that operate outside the customer organization's governed information systems, the resulting policy fragments accumulate as rule debt, a governance burden arising when control enactment occurs outside the information systems that previously made enactment governable, while a new locus of decision rights (agent instructions) emerges outside the architectural and process control points the literature has previously analyzed.

The theory also offers an IS design implication. Agentic systems should not treat prompts, tool descriptors, workflow traces, evidence records, and signoff states as peripheral metadata. When these artifacts encode rules or support accountable use, they are part of the governance architecture of the information system. This point links the theory to design questions that are central to IS theory development: what features must agentic information systems include so that modular execution does not detach from organizational accountability? The theory predicts that registries for agent instructions, versioned rule repositories, test harnesses, behavior logs, lineage records, and exception ownership are not merely implementation details; they are mechanisms through which organizations keep the accountability boundary aligned with the execution boundary.

Third, the theory extends digital platform governance theory by adding accountability-asset cospecialization to the list of structural determinants of complementor vulnerability \citep{tiwana2010,constantinides2018,parker2017}. Complementors whose capabilities are cospecialized with accountability assets are less vulnerable to envelopment than complementors whose capabilities lack such cospecialization, because the orchestrator must reproduce not only the task output but also the defensibility infrastructure that makes the output usable. The cross-sectional variation in complementor vulnerability is partly explained by the cospecialization of accountability assets with the focal capability, a refinement with implications for the broader literature on power asymmetries in digital ecosystems \citep{mcintyre2017}.

\subsection{Implications for Firm-Boundary, Complementary-Assets, and Appropriability Theory}

Beyond its IS contributions, the paper extends firm-boundary theory by identifying a boundary that can diverge from the technical interface. Coasian logic correctly predicts that agentic orchestration lowers interface and assembly costs and increases pressure toward disaggregation. The theory developed here specifies why that pressure is conditional: accountability boundaries in agentic ecosystems depend not only on the cost of assembling tasks, but also on the cost of verifying outputs for accountable use and the transferability of responsibility. The contribution also clarifies the level of analysis. The firm is often too coarse for studying AI-enabled boundary strategy. A single firm may contain capabilities with different verification costs, responsibility regimes, and accountability assets. Treating the AI-enabled capability and the offering portfolio together explains how vertical firms can simultaneously sell componentized edge functions and retain integrated accountability around the core service.

The paper also extends Teece's complementary-assets framework by specifying accountability assets as a theoretically distinct subset of complementary assets whose value often derives from cospecialization with the focal AI-enabled capability. Traditional complementary assets help commercialize an innovation; accountability assets make outputs defensible and assignable after the fact. The value of an AI capability depends on the asset that makes it legitimate, not only on the asset that distributes or services it. The argument refines appropriability logic for AI: when capabilities are imitable, firms may not capture durable value from model output alone, and value shifts toward cospecialized assets that are harder to reproduce, including evidence systems, review histories, signoff authority, governed workflows, and customer trust.

\subsection{Generalization Beyond Agentic AI}

The mechanism developed in this paper generalizes beyond agentic AI to settings in which modularization pressure meets accountability constraints. Wherever responsibility for outputs is concentrated in cospecialized assets that are costly to reproduce at the interface, the predictions should apply. Platform-mediated professional services, regulated digital services, and knowledge-intensive work organized around defensible decisions all exhibit this structure. The framework does not extend straightforwardly to settings dominated by physical-asset specificity, commodity goods, or activities where accountability for outputs is institutionally light; where defensibility is not part of value, the prior frameworks remain adequate.

\subsection{Managerial Implications}

Managers should decompose by accountability rather than interface. The first diagnostic question is not whether an agent can call the capability but whether the output can be cheaply verified for accountable use and whether responsibility can transfer. If both conditions hold, a component strategy is attractive. If either condition fails, managers should identify which accountability assets must remain under firm, customer, or professional control. Vertical AI firms should make expertise easy to call and accountability hard to copy: sell componentizable edge functions where accountability is light, retain evidence trails, governed workflows, review processes, permissions, evaluation records, authoritative repositories, and signoff where accountability is heavy, and avoid excessive dependence on a single orchestrator. Customer organizations should treat agent instructions as potential policy artifacts. Practitioner research echoes this view, arguing that scaling AI agents successfully requires treating them as team members subject to onboarding, role definition, supervision, and accountability rather than as autonomous tools \citep{telang2026}. When an instruction tells an agent how to classify, approve, escalate, summarize, or decide, the instruction should have an owner, version history, test evidence, monitoring, and reconciliation with formal policy. Without this governance, the customer accumulates rule debt. The recommendation aligns with the IT governance literature's emphasis on explicit decision rights \citep{weill2004,tiwana2014}, extended to a setting in which the rule-bearing artifacts now include natural-language agent instructions.

\section{LIMITATIONS AND FUTURE RESEARCH}

This paper develops a conceptual theory and supports it with structured theoretical illustrations. Several limitations bound the contribution and identify productive directions for future work. The most important limitation is that the theory identifies a nomological structure for accountability-boundary strategy, but it does not yet provide a fully specified design theory for building accountable agentic systems. A design-science extension could translate the theory into design principles, artifact requirements, and evaluation criteria for rule registries, agent-control dashboards, lineage systems, and signoff infrastructure.

First, the propositions are not subjected to empirical test. The structured theoretical illustrations are selected to discipline the boundary strategy logic, not to provide empirical evidence. Future research should test the propositions across larger samples of AI-enabled capabilities through comparative case research, survey research, archival research using regulatory filings or product-launch data, and quasi-experimental designs that exploit institutional shocks (regulatory changes, certification regimes, liability rulings). To facilitate this work, Table 5 provides suggested operationalizations of the core constructs, offering pathways for both qualitative case research and quantitative archival studies.

\begin{table}[ht!]
\centering
\caption{Suggested Empirical Operationalization of Core Constructs}\label{tab:operationalizations}
\renewcommand{\baselinestretch}{1.0}\normalsize
\footnotesize
\begin{tabularx}{\textwidth}{p{1.6in} X}
\toprule
\textbf{Construct} & \textbf{Suggested measurement strategies} \\
\midrule
Interface-and-assembly cost reduction & Quantitative: reduction in integration time, number of manual tool-selection or sequencing steps eliminated, decrease in workflow assembly cost, reduction in average time from task initiation to composed output after orchestrator adoption. Qualitative: customer reports that external capabilities can be discovered, invoked, and composed without custom integration work. \\
\addlinespace
Verification cost & Quantitative: average expert time required to verify whether an AI output is correct, complete, defensible, and usable in organizational action, including evidence assembly where needed. Qualitative: credential level of the required human-in-the-loop reviewer (e.g., junior analyst, senior partner, licensed professional). \\
\addlinespace
Responsibility transferability & Contract analysis (presence and scope of indemnification clauses, service-level warranties, liability caps); regulatory mapping (e.g., strict adherence to FDA non-device CDS criteria versus unregulated software); professional-rule mapping (e.g., ABA Formal Opinion 512 applicability). \\
\addlinespace
Accountability-asset cospecialization & Integration-depth analysis: the share of accountability infrastructure (audit logs, review workflows, signoff procedures, evaluation records, exception ownership) that must be tailored to the focal capability rather than reused across generic models or capabilities; supplementary indicators include the extent to which the capability's outputs cannot be defended without the specific accountability infrastructure surrounding them, and the share of off-the-shelf audit and review tools that fail to provide equivalent defensibility for the focal capability. \\
\addlinespace
Boundary strategy & Coded from product documentation, API references, regulatory filings, and customer disclosures as component, integrated, or dual-track. Secondary implementation code: fully internal execution versus bounded external computation under retained accountability control. \\
\addlinespace
Value capture & Quantitative: price premium for accountability-enabled versions of the capability; renewal or retention rates for integrated versus component offerings; gross margin by boundary strategy; share of customer spend retained after orchestrator integration. Qualitative: customer willingness to pay for evidence, review, signoff, and governed workflow features relative to raw task execution. \\
\addlinespace
Orchestrator dependence and intent capture & Share of capability invocations originating from a single orchestrator; share of user goal formulation occurring inside the orchestrator versus the complementor's surface; frequency with which the orchestrator records intermediate user feedback, alternative-tool selections, and goal refinements that the complementor does not see. \\
\addlinespace
Rule debt accumulation & IT service management metrics: volume of undocumented agent prompts and tool descriptors discovered during compliance audits; frequency of policy-reconciliation events; count of agent-driven decisions lacking a formal policy owner, version history, or test record. \\
\bottomrule
\end{tabularx}
\normalsize

\vspace{0.5em}
{\scriptsize\textit{Note. Operationalizations are illustrative rather than exhaustive. Researchers should expect to combine multiple proxies, particularly for accountability-asset cospecialization and rule debt, where single-source measurement is unlikely to be sufficient.}\par}
\renewcommand{\baselinestretch}{2.0}\normalsize
\end{table}

Second, the theory focuses on AI-enabled capabilities whose outputs are used in accountable organizational action. It does not extend straightforwardly to settings dominated by commodity exchange or conventional asset-specific investment unrelated to accountable AI-supported outputs. Some physical domains, such as pharmaceutical supply chains, industrial maintenance, autonomous systems, and regulated manufacturing, may fall within scope when AI-supported outputs must be evidenced, reviewed, certified, or assigned to an accountable party. The relevant boundary condition is not whether the domain is physical or digital, but whether accountability for the output is institutionally meaningful.

Third, the dynamic predictions of the theory require longitudinal study. The boundary migration argument predicts that capabilities move toward components as assurance institutions develop and toward integration as failures reveal hidden verification costs or as rule debt accumulates. These trajectories unfold over years, and the lag between boundary choice and emergent governance cost makes the dynamics empirically challenging. Industry studies that track specific capabilities across multiple regulatory and technological cycles, including robo-advisory, automated underwriting, algorithmic content moderation, and AI-assisted legal services, would be especially valuable.

Fourth, the framework treats verification cost and responsibility transferability as exogenous in the static analysis, but both are endogenous in the longer run. Firms can invest in reducing verification cost through evaluation tooling, traceability, and provenance infrastructure; industries can develop institutions that increase responsibility transferability. The framework also takes the orchestrator's behavior as given. Strategic decisions of orchestrators, including how aggressively to envelope adjacent functionality, how to manage complementors, and how to allocate demand, are not modeled explicitly. Future research could model these as second-order strategic moves and could use game-theoretic or formal analytical treatments to endogenize orchestrator strategy and characterize equilibrium boundary strategies under different orchestrator behaviors.

Fifth, the framework does not address heterogeneity in customer willingness to pay for accountability. Some customers are highly sensitive to defensibility, including regulated enterprises, public-sector buyers, and institutions exposed to litigation, while others are not. Variation in customer-side accountability requirements may produce variation in boundary strategies across segments of the same market. Future research should examine the demand-side heterogeneity that this paper has held implicit.

Sixth, rule debt warrants focused investigation within IS control and IT governance research. Productive directions include studying how organizations discover and pay down rule debt, whether providers that supply governance tooling capture value as predicted, and how rule debt relates to technical debt, shadow IT, and control configuration in IS projects.

\section{CONCLUSION}

Agentic AI orchestrators reduce the interface and assembly costs that historically favored integration in knowledge-intensive work, but lower assembly costs do not automatically produce vertical disintegration. When AI-enabled outputs require defensibility, reviewability, signoff, and assignable responsibility, the strategically relevant assets are accountability assets, not merely interface assets. These assets shape boundary choice, value capture, and investment incentives. Boundary misconfiguration creates rule debt when decision rules migrate into ungoverned agentic execution environments without ownership, versioning, testing, and reconciliation. Agentic AI will not simply dissolve organizational boundaries; it will redraw them around accountability, and the information systems that govern accountability will become central to where they settle.

\renewcommand{\baselinestretch}{1.0}\normalsize
\setlength{\bibsep}{3pt}
\bibliographystyle{apalike}
\bibliography{_manual_misq_vAI_references_v14}
\renewcommand{\baselinestretch}{2.0}\normalsize

\clearpage
\renewcommand{\baselinestretch}{1.0}\footnotesize   
\section*{Online Supplementary Material}
\addcontentsline{toc}{section}{Online Supplementary Material}

\setcounter{subsection}{0}
\renewcommand{\thesubsection}{\Alph{subsection}}

\begin{center}{\large\textbf{Online Supplementary Material for ``Redrawing the AI Map''}}\end{center}
\vspace{0.5em}

	\textbf{Note.} This online supplementary material is not necessary to understand the paper. All primary arguments, constructs, propositions, illustrations, and claims are contained in the main manuscript. The manuscript does not rely on or reference materials in this document, although this document may refer back to relevant parts of the manuscript. The material records illustrative content that clarifies how the theory can be read across use cases, how the three boundary strategies differ visually, and how modular access to an accountability-bearing core differs from componentization of the accountability boundary. It is illustrative rather than evidentiary.

\begin{landscape}
\begin{table}[p]
\centering
\caption{Construct-by-Use-Case Crosswalk}
\scriptsize
\setlength{\tabcolsep}{1pt}
\begin{tabularx}{\linewidth}{P{1.16in} P{1.55in} P{1.49in} P{1.74in} P{1.58in} P{1.38in}}
\toprule
\textbf{Use case / capability} & \textbf{Verification cost} & \textbf{Responsibility transferability} & \textbf{Accountability assets} & \textbf{Rule debt risk} & \textbf{Predicted boundary strategy} \\
\midrule
Document extraction and classification & Low. Outputs can be compared to source documents, schemas, deterministic checks, or statistical samples. & High. Accuracy can often be governed by service-level commitments, warranties, reprocessing, or customer review. & Source linkage, extraction logs, field-level confidence, schema validation, and exception queues. Cospecialization is usually limited unless extraction directly supports a regulated conclusion. & Low to moderate. Risk rises when extraction rules, thresholds, or exception policies migrate into unmanaged prompts. & Component. The capability can be sold or invoked as a callable function. \\
\addlinespace
Legal research retrieval & Low for the retrieval output. Existence of authorities, source passages, and coverage can be checked. & High for the retrieval output. Providers can warrant search coverage, citation accuracy, provenance, and database reliability. & Source linkage, provenance logs, citator metadata, confidentiality controls, and retrieval audit logs. Downstream legal advice is a separate accountable capability. & Moderate. Risk arises if research instructions encode legal standards or matter-specific policies outside governed systems. & Component. Downstream professional use is governed, but external to the retrieval-capability boundary. \\
\addlinespace
Audit evidence and signoff & High. The conclusion depends on evidence quality, materiality, review history, professional judgment, and support for conclusions. & Low. Audit responsibility and signoff cannot be transferred to a general orchestrator or external component. & Evidence trails, workpapers, materiality judgments, review workflows, lineage records, permissions, and signoff authority. & High. Rule debt accumulates if audit procedures, sampling rules, or exception logic migrate into unmanaged agent instructions. & Integrated. Bounded computational services may be used only inside controlled accountability infrastructure. \\
\addlinespace
Clinical decision support and care planning & High. Clinical use depends on patient context, provenance, escalation, clinician judgment, and documentation. & Low. The clinician or care organization retains responsibility for diagnosis, treatment, escalation, and follow-up. & Patient record integration, clinical protocols, evidence provenance, escalation paths, clinician review, and care-plan documentation. & High. Risk arises if triage rules, care protocols, or escalation standards become unversioned agent instructions. & Integrated. Modular front doors may initiate work, but clinical accountability remains governed. \\
\addlinespace
Legal drafting versus filed opinion & Mixed. Drafts can be reviewed and edited; filed opinions or court submissions require defensible professional commitment. & Mixed. Draft support can be used as an input, but the filed commitment remains with the lawyer or firm. & For drafting: source references and comparison logs. For filed commitments: client record, review history, professional signoff, and tribunal-facing evidence. & Moderate to high. Risk rises when matter-specific legal standards or filing rules are embedded in unmanaged prompts. & Dual-track. Drafting and comparison can be components; filed commitments remain integrated. \\
\addlinespace
Procurement scoring versus supplier approval & Mixed. Scores can be checked against inputs and models; approvals require policy compliance, exception handling, and accountable ownership. & Mixed. Component providers can support scoring, but the customer retains approval responsibility. & Risk-score provenance, supplier records, policy repositories, exception logs, committee approvals, and contract records. & High when supplier rules, thresholds, and exception policies migrate from governed procurement systems into prompts. & Dual-track. Risk scoring can be componentized; supplier approval remains governed. \\
\bottomrule
\end{tabularx}
\end{table}
\end{landscape}

\clearpage
\begin{landscape}
\begin{figure}[p]
\centering
\begin{tikzpicture}[
  font=\sffamily\small,
  >=Stealth,
  user/.style={circle,draw=black!80,fill=gray!18,very thick,minimum size=1.15cm,align=center},
  orch/.style={rectangle,draw=blue!80,fill=blue!10,very thick,rounded corners,minimum height=1.0cm,minimum width=2.6cm,align=center},
  comp/.style={rectangle,draw=gray!80,fill=gray!10,thick,dashed,minimum height=1.0cm,minimum width=2.8cm,align=center},
  flow/.style={->,thick,draw=black!75}
]
\node[font=\sffamily\bfseries\Large] at (6.4,2.1) {Component strategy};
\node[user] (u1) at (0,0) {User};
\node[orch] (o1) at (3.0,0) {Agentic\\orchestrator};
\node[comp] (c1) at (6.8,0) {Vertical AI\\component};
\node[user,minimum size=1.55cm] (cu1) at (10.8,0) {Customer\\use};
\draw[flow] (u1) -- (o1);
\draw[flow] (o1) -- node[above]{call} (c1);
\draw[flow] (c1) -- node[above,align=center]{component\\output} (cu1);
\end{tikzpicture}
\caption{Boundary Strategies in Agentic Ecosystems: Component Strategy. The capability or sub-capability is sold or invoked as a callable function. Verification is low, or assurance makes responsibility transferable.}
\end{figure}
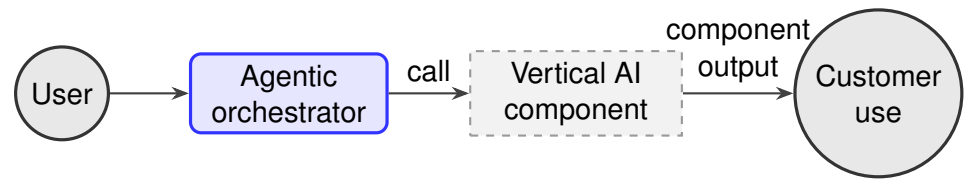
\end{landscape}

\clearpage
\begin{landscape}
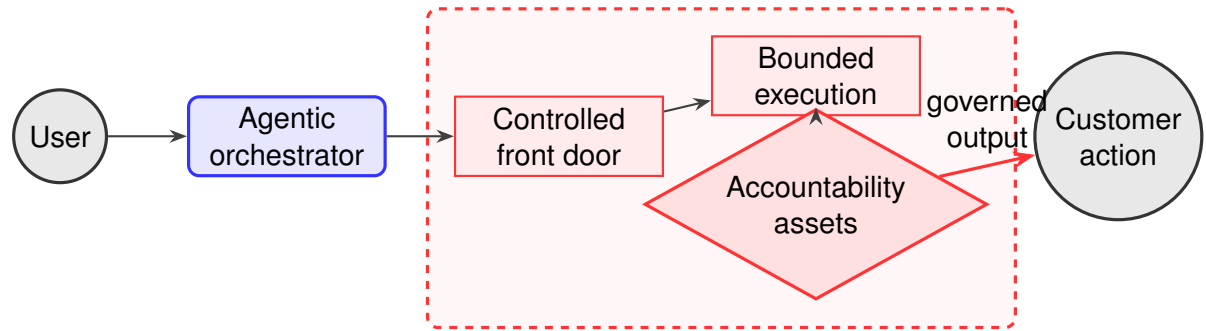
\begin{figure}[p]
\centering
\begin{tikzpicture}[
  font=\sffamily\small,
  >=Stealth,
  user/.style={circle,draw=black!80,fill=gray!18,very thick,minimum size=1.15cm,align=center},
  orch/.style={rectangle,draw=blue!80,fill=blue!10,very thick,rounded corners,minimum height=1.0cm,minimum width=2.6cm,align=center},
  vfirm/.style={rectangle,draw=red!80,fill=red!8,thick,minimum height=1.0cm,minimum width=2.75cm,align=center},
  gate/.style={diamond,draw=red!80,fill=red!12,very thick,aspect=1.8,minimum width=2.75cm,align=center},
  boundary/.style={draw=red!80,fill=red!3,very thick,dashed,rounded corners,inner sep=0.35cm},
  flow/.style={->,thick,draw=black!75},
  acct/.style={->,very thick,draw=red!80}
]
\node[font=\sffamily\bfseries\Large] at (7.0,2.5) {Integrated strategy};
\node[user] (u2) at (0,0) {User};
\node[orch] (o2) at (3.0,0) {Agentic\\orchestrator};
\node[vfirm] (front2) at (6.6,0) {Controlled\\front door};
\node[vfirm] (exec2) at (10.0,0.8) {Bounded\\execution};
\node[gate] (gate2) at (10.0,-0.9) {Accountability\\assets};
\node[user,minimum size=1.55cm] (cu2) at (14.0,0) {Customer\\action};
\begin{scope}[on background layer]
  \node[boundary,fit=(front2) (exec2) (gate2)] (b2) {};
\end{scope}
\draw[flow] (u2) -- (o2);
\draw[flow] (o2) -- (front2);
\draw[flow] (front2) -- (exec2);
\draw[flow] (exec2) -- (gate2);
\draw[acct] (gate2) -- node[above,align=center]{governed\\output} (cu2);
\end{tikzpicture}
\caption{Boundary Strategies in Agentic Ecosystems: Integrated Strategy. A modular front door may initiate work or return signed outputs, but evidence, review, lineage, signoff, and responsibility remain inside the retained accountability boundary.}
\end{figure}
\end{landscape}

\clearpage
\begin{landscape}
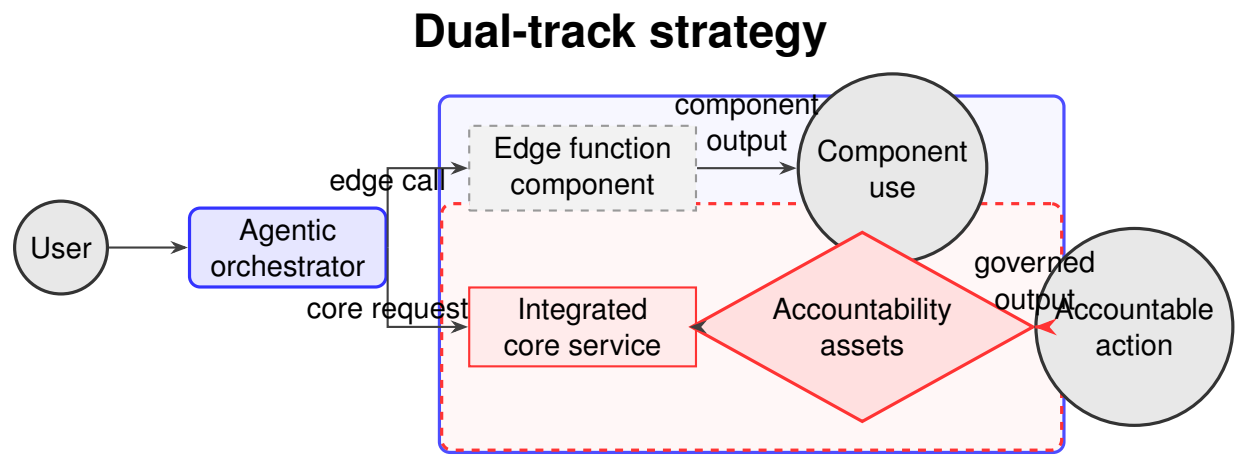
\begin{figure}[p]
\centering
\begin{tikzpicture}[
  font=\sffamily\small,
  >=Stealth,
  user/.style={circle,draw=black!80,fill=gray!18,very thick,minimum size=1.15cm,align=center},
  orch/.style={rectangle,draw=blue!80,fill=blue!10,very thick,rounded corners,minimum height=1.0cm,minimum width=2.6cm,align=center},
  comp/.style={rectangle,draw=gray!80,fill=gray!10,thick,dashed,minimum height=1.0cm,minimum width=3.0cm,align=center},
  vfirm/.style={rectangle,draw=red!80,fill=red!8,thick,minimum height=1.0cm,minimum width=3.0cm,align=center},
  gate/.style={diamond,draw=red!80,fill=red!12,very thick,aspect=1.8,minimum width=2.75cm,align=center},
  boundary/.style={draw=red!80,fill=red!3,very thick,dashed,rounded corners,inner sep=0.35cm},
  offer/.style={draw=blue!70,fill=blue!3,very thick,rounded corners,inner sep=0.38cm},
  flow/.style={->,thick,draw=black!75},
  acct/.style={->,very thick,draw=red!80}
]
\node[font=\sffamily\bfseries\Large] at (7.4,2.8) {Dual-track strategy};
\node[user] (u3) at (0,0) {User};
\node[orch] (o3) at (3.0,0) {Agentic\\orchestrator};
\node[comp] (edge3) at (6.9,1.05) {Edge function\\component};
\node[user,minimum size=1.4cm] (cu3a) at (11.0,1.05) {Component\\use};
\node[vfirm] (core3) at (6.9,-1.05) {Integrated\\core service};
\node[gate] (gate3) at (10.6,-1.05) {Accountability\\assets};
\node[user,minimum size=1.55cm] (cu3b) at (14.2,-1.05) {Accountable\\action};
\begin{scope}[on background layer]
  \node[offer,fit=(edge3) (core3) (gate3)] (offer3) {};
  \node[boundary,fit=(core3) (gate3)] (b3) {};
\end{scope}
\draw[flow] (u3) -- (o3);
\draw[flow] (o3.east) |- node[near start,above,align=center]{edge call} (edge3.west);
\draw[flow] (edge3) -- node[above,align=center]{component\\output} (cu3a);
\draw[flow] (o3.east) |- node[near start,below,align=center]{core request} (core3.west);
\draw[flow] (core3) -- (gate3);
\draw[acct] (gate3) -- node[above,align=center]{governed\\output} (cu3b);
\end{tikzpicture}
\caption{Boundary Strategies in Agentic Ecosystems: Dual-Track Strategy. Separable edge functions are componentized while the accountability-bearing core remains integrated. Dual-track separates functions, not one accountability-bearing output split between external execution and internal signoff.}
\end{figure}
\end{landscape}

\clearpage
\subsection{Bounded Modular Access to an Accountability-Bearing Core}
\label{app:bounded-modular-access}

This section supplements the integrated and dual-track strategy discussions in Section~\ref{sec:integrated-strategy} and Section~\ref{sec:dual-track-strategy}. The accountability-bearing core can have a modular front door without becoming a modular accountability boundary. A vertical AI firm can expose an API, tool endpoint, or agent-callable interface for initiating work, submitting inputs, checking status, or retrieving governed outputs. The accountability-bearing work remains integrated when the firm retains control over the review workflow, evidence trail, lineage, signoff process, system of record, and final commitment.

For an accountability-bearing core, modularity can exist at the level of access protocol, not at the level of responsibility transfer. Table~\ref{tab:bounded-modular-access} gives illustrative interface actions.

\begin{table}[h]
\centering
\caption{Bounded Modular Access to an Integrated Accountability Core}
\label{tab:bounded-modular-access}
\small
\begin{tabular}{p{0.32\linewidth} p{0.35\linewidth} p{0.25\linewidth}}
\toprule
\textbf{Interface action} & \textbf{What it does} & \textbf{Boundary implication} \\
\midrule
\texttt{submit\_case()} &
The orchestrator submits facts, documents, or task parameters. &
The interface is modular. \\
\texttt{request\_review()} &
The orchestrator asks the vertical firm to perform accountable review. &
The interface is modular. \\
\texttt{retrieve\_status()} &
The orchestrator checks whether review is pending, completed, or escalated. &
The interface is modular and may be asynchronous. \\
\texttt{retrieve\_signed\_output()} &
The orchestrator receives the final governed output after signoff. &
Accountability remains retained. \\
\texttt{retrieve\_evidence\_packet()} &
The customer receives a controlled evidence record. &
Accountability remains retained if evidence generation and custody remain firm-controlled. \\
\bottomrule
\end{tabular}
\end{table}

In all of these cases, the orchestrator can interact with the service through a modular interface. But it does not get to produce the accountable conclusion, control the signoff workflow, modify the evidence trail, or own the accountability-relevant traces. The firm can let the orchestrator initiate a governed review; it should not let the orchestrator generate the accountable conclusion, assemble the evidence, decide whether the output is defensible, and present it as final.

The clean distinction is therefore between the interface layer, execution layer, and accountability layer. The interface layer can be modular because agents can initiate, route, submit inputs, retrieve status, and receive outputs through standardized protocols. The execution layer can sometimes rely on bounded external computation if doing so does not expose or transfer accountability-relevant control. The accountability layer usually remains integrated because evidence, review, lineage, signoff, and responsibility must remain controlled by the vertical firm or regulated accountability holder.

\clearpage
\subsection{How Accountability Boundary Shifts Occur}
\label{app:accountability-boundary-shifts}

This section supplements the main manuscript's discussion of orchestrator dependence, intent capture, and platform appropriation in Section~\ref{sec:orchestrator-intent-capture}, Section~\ref{sec:appropriability}, and Figure~\ref{fig:theoretical-model}. A shift in the accountability boundary does not require a formal transfer of legal responsibility or ownership of data. The vertical AI firm may remain contractually, professionally, or reputationally exposed even while the practical control needed to discharge accountability moves elsewhere. The mechanism is architectural and operational: evidence, review, lineage, signoff, logs, feedback, and responsibility records begin to be created, stored, interpreted, or controlled outside the vertical firm's governed system, often at the orchestrator interface.

Table~\ref{tab:formal-accountability-boundary} distinguishes formal accountability from the accountability boundary.

\begin{table}[h]
\centering
\caption{Formal Accountability versus the Accountability Boundary}
\label{tab:formal-accountability-boundary}
\small
\begin{tabular}{p{0.24\linewidth} p{0.68\linewidth}}
\toprule
\textbf{Concept} & \textbf{Meaning} \\
\midrule
Formal accountability &
Who is legally, contractually, professionally, or reputationally responsible after the fact. \\
Accountability boundary &
Who controls the assets and processes that make an output defensible, reviewable, auditable, and assignable. \\
\bottomrule
\end{tabular}
\end{table}

The distinction matters because formal accountability and practical control can separate. For example, suppose an audit AI firm lets an orchestrator handle the user workflow. The orchestrator collects the client request, chooses what evidence to send, frames the task, stores the interaction history, presents the output, captures the user's acceptance, and logs the final workflow. The audit AI firm may still generate a conclusion, and it may still be blamed or contracted against if the conclusion fails. Yet the accountability-relevant record is now partly controlled by the orchestrator. The firm no longer controls the full evidentiary and review context needed to justify, review, and discharge accountability.

Table~\ref{tab:accountability-boundary-shift} identifies common mechanisms through which this shift can occur.

\begin{table}[h]
\centering
\caption{Mechanisms of Accountability Boundary Shift}
\label{tab:accountability-boundary-shift}
\small
\begin{tabular}{p{0.30\linewidth} p{0.62\linewidth}}
\toprule
\textbf{Mechanism} & \textbf{What shifts} \\
\midrule
The orchestrator frames the task &
The orchestrator controls what problem was asked, what facts were included, and what constraints mattered. \\
The orchestrator collects or filters evidence &
The vertical firm no longer controls the evidentiary basis for the output. \\
Review occurs in the orchestrator workspace &
Review comments, edits, approvals, escalations, and exceptions are stored outside the vertical firm's governed system. \\
Signoff occurs through the orchestrator interface &
The final commitment is recorded at the interface rather than inside the governed workflow. \\
The orchestrator stores interaction history &
The orchestrator observes which outputs users accept, reject, revise, or escalate. \\
The orchestrator controls routing and ranking &
The vertical firm loses control over when and how its capability is invoked. \\
The customer relies on orchestrator logs &
The practical record of responsibility sits outside the vertical firm's accountability infrastructure. \\
\bottomrule
\end{tabular}
\end{table}

The same contrast can be stated as a design principle. In the problematic design, the orchestrator gathers documents, decides what evidence matters, asks the vertical AI firm for a conclusion, presents the result to the user, captures the review trail, and records user approval. The vertical AI firm still supplies an underlying capability, but the accountability-relevant assets are created and controlled around the orchestrator. In the safer design, the orchestrator can initiate the request and retrieve the final governed output, but the vertical AI firm controls evidence collection, review, escalation, signoff, lineage, and the final defensible record.

Thus, making a capability callable is not equivalent to transferring the accountability boundary. Callable interfaces can expand markets when they allow agents to initiate, route, and retrieve work. They become strategically dangerous when they allow evidence, review, signoff, lineage, or responsibility records to be created or controlled outside the firm's governed system. The design problem for vertical AI firms is therefore to expose bounded access points without ceding control over the assets needed to justify, review, and discharge accountability.

\clearpage
\subsection{Why Component Strategies Persist When Frontier Models Improve}
\label{app:components-frontier-models}

This section supplements the component strategy logic in Section~\ref{sec:component-strategy} and Table~\ref{tab:boundary-strategies}. A natural objection to the component strategy is that frontier models may eventually perform many bounded enterprise tasks without specialized external tools. The relevant comparison, however, is not whether a frontier model can sometimes perform a task in isolation. The comparison is whether an enterprise workflow can rely on the model alone at the required level of accuracy, latency, cost, schema fidelity, security, operational traceability, and integration with downstream systems. In many settings, specialized components remain valuable because they provide bounded functionality with stronger operational guarantees than a general-purpose model call.

Document intelligence illustrates the point. In document-intensive workflows, the bottleneck is often not reasoning over clean inputs but reliably reading, parsing, locating, and structuring messy enterprise documents before reasoning begins.\footnote{See Archika Dogra, Sergei Tsarev, and Erich Elsen, ``Why Your Agents Can't Read Enterprise Documents,'' Databricks Blog, \url{https://perma.cc/2L5L-2B56}; and the Databricks AI Research Team, ``Introducing OfficeQA: A Benchmark for End-to-End Grounded Reasoning,'' \url{https://perma.cc/9W2C-JKWQ}. These sources report that frontier agents struggle on real-world enterprise documents, especially scanned PDFs, nested tables, handwritten notes, inconsistent layouts, and multi-document reasoning tasks.} A frontier model may serve as the orchestrator or reasoning layer, while specialized components perform bounded tasks with stronger guarantees. Document-intelligence tools can parse complex layouts once into reusable structured representations; retrieval systems can narrow the search space before expensive model calls; classification models can route work cheaply; and formatting tools can enforce schemas that downstream systems require. The component provider therefore does not compete only on generic intelligence. It competes on accuracy, reliability, availability, latency, price, security, schema discipline, and ease of integration.

\begin{table}[h]
\centering
\caption{Why Specialized Components Remain Useful Alongside Frontier Models}
\label{tab:frontier-model-components}
\small
\begin{tabular}{p{0.22\linewidth} p{0.34\linewidth} p{0.34\linewidth}}
\toprule
\textbf{Task} & \textbf{Why frontier models alone may be insufficient} & \textbf{Component advantage} \\
\midrule
Data extraction from standardized documents &
Scanned PDFs, nested tables, vendor-specific forms, handwriting, and layout variation can produce extraction errors that contaminate downstream reasoning. &
Layout-aware parsers, OCR pipelines, and document-intelligence components can provide structured fields, source links, confidence scores, and repeatable extraction behavior. \\
Document classification and first-pass retrieval &
Using a frontier model for every classification or retrieval step can be slow and costly, especially at high volume. &
Lightweight classifiers, indexes, and vector-search systems can route documents and narrow context before the frontier model is invoked. \\
Simple formatting and schema conversion &
Generative models may produce subtle deviations from required JSON, XML, or tabular schemas. &
Programmatic validators, parsers, and task-specific models can enforce deterministic schemas needed by downstream systems. \\
Low-stakes summarization &
Routine summarization may not require frontier-level reasoning, and high-volume use can create unnecessary cost and latency. &
Smaller or task-specific summarization components can provide adequate quality at lower cost and with more predictable throughput. \\
\bottomrule
\end{tabular}
\end{table}

The implication is that the component strategy does not assume frontier models are weak. It assumes that, for some capabilities, the output is sufficiently bounded, verifiable, and responsibility-transferable that a specialized component can be invoked as a market service. In such cases, the orchestrator benefits from calling the component rather than reproducing it, and the customer can cheaply evaluate the output or shift responsibility through contract or routine operational processes. The component strategy is therefore most plausible where specialized tools provide operational reliability that frontier models alone do not yet deliver, and where the task does not require the provider to retain an accountability-bearing core.

\clearpage
\subsection{How Accountability-Bearing Capabilities Can Migrate Toward Components}
\label{app:migration-to-components}

This section supplements the manuscript's discussion of scope conditions and migration logic in Section~\ref{sec:scope-boundary-conditions}, as well as the dual-track strategy in Section~\ref{sec:dual-track-strategy}. The theory does not imply that today's integrated capabilities must remain integrated permanently. Some capabilities remain integrated because execution cannot yet be separated from accountability in a defensible way. If institutional infrastructure develops, including certification regimes, professional warranties, liability allocation, insurance markets, regulatory safe harbors, audit standards, and industry protocols, responsibility may become more transferable for bounded sub-capabilities. In such cases, component strategies become feasible not because accountability disappears, but because institutions preserve accountability while allowing execution to move outside the integrated provider.

Table~\ref{tab:migration-to-components} illustrates how this migration might occur across four accountability-bearing domains. The pattern is deliberately conservative: migration usually begins with bounded execution-layer sub-capabilities whose outputs can be verified, certified, insured, or contractually warranted. Final accountable commitments, such as audit opinions, treatment decisions, filed legal arguments, or supplier approvals, remain integrated unless institutions develop accepted ways to preserve responsibility outside the integrated provider.

\begin{landscape}
\begin{table}[p]
\centering
\caption{Illustrative Migration Paths from Integrated to Component Strategies}
\label{tab:migration-to-components}
\small
\begin{tabular}{p{0.14\linewidth} p{0.24\linewidth} p{0.30\linewidth} p{0.24\linewidth}}
\toprule
\textbf{Domain} & \textbf{Currently integrated core} & \textbf{Institutional infrastructure that could enable migration} & \textbf{Likely componentizable sub-capabilities} \\
\midrule
Audit &
Audit conclusions, evidence evaluation, materiality judgments, workpaper review, and partner signoff. &
Certification of AI audit procedures; standardized evidence logs; warranties for bounded extraction or matching tasks; insurance for certified tool failures; regulator-accepted documentation trails. &
Invoice-to-purchase-order matching; lease-term extraction; workpaper formatting; anomaly flagging; evidence completeness checks; standardized confirmation tracking. \\
Clinical decision support &
Diagnosis, care planning, treatment recommendations, escalation decisions, and clinician responsibility. &
Regulatory clearance for bounded AI tools; hospital governance standards; clinical guidelines for AI-supported use; malpractice coverage for certified outputs; accepted audit trails linking inputs, model versions, and clinician action. &
Medical coding; medication-interaction checks; triage support; radiology pre-screening; patient-message summarization; extraction of structured clinical facts. \\
Legal services &
Legal opinions, filed briefs, litigation strategy, client advice, and lawyer certification of filings. &
Bar guidance on approved AI use; malpractice coverage for certified legal tools; court-accepted citation audit trails; warranties for citation validity or clause extraction; professional standards for review of AI-assisted drafts. &
Citation verification; legal research retrieval; clause extraction; brief-format compliance; privilege-screening support; first-pass document comparison. \\
Procurement &
Supplier approval, exception approval, risk acceptance, and accountable sourcing decisions. &
Industry standards for supplier-risk scoring; third-party model certification; warranties for data freshness or classification accuracy; audit trails for score generation; contractual allocation for scoring errors. &
Supplier-risk scoring; sanctions-screening checks; ESG document extraction; invoice classification; contract metadata extraction; exception flagging. \\
\bottomrule
\end{tabular}
\end{table}
\end{landscape}

The examples can be read as layer-by-layer migration rather than wholesale disaggregation. In audit, AI tools may become callable for invoice matching, evidence extraction, lease-term extraction, or workpaper formatting before audit opinions or partner signoff become transferable. In clinical decision support, AI tools may become callable for triage support, coding, radiology pre-screening, drug-interaction checking, or patient-message summarization before diagnosis and care planning move outside clinician or hospital accountability. In legal services, citation checking, legal research retrieval, clause extraction, privilege screening, or brief-format compliance may become components before filed briefs, client advice, or legal opinions become transferable. In procurement, sanctions screening, supplier-risk scoring, ESG document extraction, or contract metadata extraction may become components before supplier approval or risk acceptance moves outside the organization's governed decision process.

The examples show that migration is a change in the accountability regime, not merely a technical improvement in AI performance. A bounded task becomes componentizable when its output can be verified cheaply or responsibility can be transferred through accepted institutional mechanisms. Until then, an orchestrator may be able to invoke the capability through a modular interface, but the accountability-bearing core remains integrated. This is why dual-track strategies are likely to appear before full componentization in many professional domains: edge functions migrate first, while the accountability-bearing core remains inside the integrated boundary.

\clearpage
\subsection{Clinical Decision Support and the Componentization Boundary}
\label{app:cds-componentization}

This section expands the clinical decision-support illustration in Section~\ref{sec:clinical-illustration} and the integrated-strategy logic in Section~\ref{sec:integrated-strategy}. Clinical decision support (CDS) illustrates why technical callability is not equivalent to componentization of accountability. Some CDS functions can already be componentized: medication-interaction checks, coding support, guideline retrieval, triage prompts, or patient-message summarization can often be exposed through APIs or embedded in larger workflows. These functions produce bounded outputs whose accuracy can be checked, logged, or overridden. But CDS becomes accountability-bearing when its output contributes to diagnosis, treatment planning, escalation, or other clinical commitments. In those cases, the relevant capability is not merely the generation of a recommendation; it is the production of a recommendation that can be reviewed, justified, reconciled with the patient record, acted on by a clinician, and defended after the fact.

CDS therefore often remains integrated because the accountability assets are tied to the clinical workflow. The system must know which patient data were available, which guideline or model version was used, what uncertainty or contraindications were present, who reviewed the recommendation, whether the recommendation was accepted or overridden, and how the final decision entered the medical record. If these traces sit outside the provider's governed workflow, formal responsibility may remain with the clinician or health system while the evidentiary basis for discharging that responsibility is fragmented. The accountability boundary therefore remains inside the clinical organization or the integrated CDS provider unless responsibility can be transferred through accepted clinical, regulatory, and liability mechanisms.

A common objection is that CDS systems historically existed as standalone tools, sometimes requiring clinicians to use a separate interface or ``swivel chair'' between systems. That history does not contradict the theory. A standalone interface is not the same as a component strategy. Many standalone CDS tools struggled precisely because they were poorly integrated into clinical workflow, required duplicate data entry, interrupted clinician work, or failed to create an accountability-relevant record inside the system where clinical decisions were made.\footnote{The CDS implementation literature has long emphasized that effective decision support depends on delivering the right information to the right person, in the right format, through the right channel, at the right time. See, for example, Sirajuddin et al. (2009), \url{https://perma.cc/AG93-ZKF6}; Osheroff et al. (2012), \url{https://perma.cc/VTQ4-3AV4}; and the AHRQ CDS ``Five Rights'' framing, \url{https://perma.cc/5GZ9-EFYJ}.} Such tools may have been technically separate, but they did not necessarily make responsibility transferable or support accountable clinical action. Their limited adoption is therefore consistent with the theory: modular interface separation without accountability integration creates friction rather than a durable component strategy.

CDS could migrate further toward component strategy if institutional infrastructure made responsibility more transferable for bounded outputs. Examples include regulatory clearance for narrowly specified CDS functions, hospital governance standards for model monitoring and override logging, malpractice or vendor warranties for certified recommendations, standardized audit trails linking inputs to outputs, and clinical guidelines specifying when AI-supported recommendations can be relied on. Under those conditions, selected CDS sub-capabilities may become components. For example, drug-interaction checks, structured risk scores, coding recommendations, or radiology pre-screening may become callable services with accepted assurance regimes. But diagnosis, treatment planning, escalation, and final clinical responsibility are likely to remain integrated until institutions can preserve accountability while allowing execution to move outside the clinical accountability boundary.

Thus, the theory does not predict that CDS is categorically integrated. It predicts that CDS componentization varies by the accountability regime of the sub-capability. Low-stakes, bounded, verifiable CDS outputs can become components. CDS outputs that shape clinical commitments require integration unless assurance institutions make responsibility transferable.

\end{document}